\documentclass[conference]{IEEEtran}
\IEEEoverridecommandlockouts
\usepackage{cite}
\usepackage{amsmath,amssymb,amsfonts}
\usepackage{algorithmic}
\usepackage{graphicx}
\usepackage{textcomp}
\usepackage{balance}
\usepackage{color}
\newcommand{\ru}    {\rule{0mm}{3.5mm}}

\def\BibTeX{{\rm B\kern-.05em{\sc i\kern-.025em b}\kern-.08em
    T\kern-.1667em\lower.7ex\hbox{E}\kern-.125emX}}

\begin{document}

\title{Camera-based Image Forgery Localization\\ using Convolutional Neural Networks
{\footnotesize
}
\thanks{This material is based on research sponsored by
the Air	Force Research Laboratory and the Defense Advanced Research Projects Agency under agreement number FA8750-16-2-0204.
The U.S. Government is authorized to reproduce and distribute reprints for Governmental purposes notwithstanding any copyright notation thereon.
The views and conclusions contained herein are those of the authors and should not be interpreted
as necessarily representing the official policies or endorsements, either expressed or implied,
of the Air Force Research Laboratory and the Defense Advanced Research Projects Agency or the U.S. Government.}
}

\author{\IEEEauthorblockN{Davide Cozzolino, Luisa Verdoliva}
\IEEEauthorblockA{Department of Electrical Engineering and Information Technology} 
{University Federico II of Naples, Naples, Italy} \\
{name.surname@unina.it}
}
\maketitle

\begin{abstract}
Camera fingerprints are precious tools for a number of image forensics tasks.
A well-known example is the photo response non-uniformity (PRNU) noise pattern, a powerful {\em device} fingerprint.
Here, to address the image forgery localization problem, we rely on noiseprint, a recently proposed CNN-based camera {\em model} fingerprint. The CNN is trained to minimize the distance between same-model patches, and maximize the distance otherwise.
As a result, the noiseprint accounts for model-related artifacts just like the PRNU accounts for device-related non-uniformities.
However, unlike the PRNU, it is only mildly affected by residuals of high-level scene content.
The experiments show that the proposed noiseprint-based forgery localization method improves over the PRNU-based reference.
\end{abstract}

\begin{IEEEkeywords}
Image forensics, PRNU, convolutional neural networks.
\end{IEEEkeywords}

\section{Introduction}

With the widespread diffusion of powerful media editing tools,
falsifying images and videos has become easier and easier in the last few years.
Manipulated visual content, often used to support fake news, represents a growing menace in many fields of life,
from politics, to journalism, to the judiciary.
In response to this threat,
a large number of methods have been recently proposed for image forgery detection and localization \cite{Korus2017_review}.

Supervised methods which exploit
the presence of camera artifacts in the image under analysis \cite{Verdoliva2014,Bondi2017},
especially those relying on the PRNU pattern \cite{Lukas2006},
have shown great potential for many forensic tasks.
The PRNU, caused by sensor defects arising in the manufacturing process,
links univocally each photo to the device that acquired it, and can be therefore regarded as a sort of device fingerprint.
PRNU-based methods have shown a very good performance for both
source identification and image forgery detection \cite{Lukas2006, Chen2008, Chierchia2014, Korus2016, Chakraborty2017, Korus2017}.
The camera PRNU pattern must be accurately estimated in advance,
which requires the availability of the camera itself or of a certain number of photos (typically, 100-200) taken from it.
At testing time, this reference pattern is compared with the single-image PRNU estimate extracted from the image under analysis.
For camera identification, the comparison takes place on the whole image.
Instead, for forgery detection and localization, a sliding-window correlation-based procedure is used.
In the presence of a forgery, the reference PRNU is missing, and a low correlation is observed.

Ideally, this procedure allows one to detect and localize accurately all attacks to the image under test.
In practice, the PRNU traces found in any individual image
represent a very weak signal in strong noise (scene residuals, camera artifacts) which makes the whole process quite unreliable.
To improve the signal-to-noise ratio,
the high-level scene content (seen as noise in this context) is removed my means of suitable denoising filters,
obtaining a noise residual which represents the desired single-image PRNU estimate.
However, due to imperfections of denoising filters,
some scene contents leak in the noise residual, inducing false alarms in dark, uniform, or very textured areas.
Besides using state-of-the-art denoising filters \cite{Chierchia2010,Al-Ani2017},
a number of strategies have been proposed to face this problem.
In \cite{Chen2008} a predictor is used to identify potentially troubling regions and adapt the statistical test locally,
while in \cite{Li2010} the interference of scene details is reduced by selective attenuation of wavelet coefficients.

A further source of noise, besides scene residuals, is represented by the so-called non-unique artifacts \cite{Fridrich2012},
traces of in-camera processes which are specific of a camera model but not of the individual device.
Such artifacts are caused, for example, by JPEG compression or CFA interpolation, and are characterized by spatially periodic patterns.
Again, various strategies have been proposed to remove them \cite{Chen2008, Li2012}.

However, one should remember that estimating the PRNU is not a goal in itself, but a means towards the completion of forensic tasks.
Do model-related artifacts really hamper such tasks? Or else can they be exploited to improve performance?
Under an information-theoretic point of view, the answer is obvious: all available information should be taken into account.
Based on this line of thought,
in \cite{Cozzolino_noiseprint} we proposed a new approach to identify the camera model traces and exploit them for multimedia forensics.
Unlike in the prior literature, which focuses on compact features,
we extract a PRNU-like camera model fingerprint, called {\em noiseprint},
which displays non-unique artifacts in the form of an image-size pattern (see Fig.1).
The scene content is effectively removed by means of a siamese residual-based convolutional neural network,
obtaining eventually a strong pattern affected by weak noise,
which allows the easy (even visual) detection and localization of anomalies due to local image manipulations.

In \cite{Cozzolino_noiseprint} we developed a totally {\em blind} noiseprint-based forgery localization technique,
working only on the image residual (single-image noiseprint estimate) with no side information.
Here, we consider the same problem in a supervised setting.
Given a suitable training set of images, a reliable estimate of the noiseprint is built,
and used in a PRNU-like fashion to discover anomalies in the residual of the image under analysis.
In fact, when a region of the image is tampered with, its noiseprint is perturbed,
that is, replaced with a new one (splicing), strongly modified (rotation, resizing), or even deleted (inpainting),
which allows one to detect and localize the attack.
It is worth reminding that the noiseprint contains only weak traces of the PRNU itself, hence it is not device-specific.
Therefore, it should not be regarded as a substitute of the PRNU.
On the other hand, the noiseprint has a much higher signal-to-noise ratio than the PRNU pattern,
a property which guarantees a better performance and allows its applications to more challenging situations.

\begin{figure}[b!]
	\centering
	\begin{tabular}{cc}
		\includegraphics[scale=0.11]{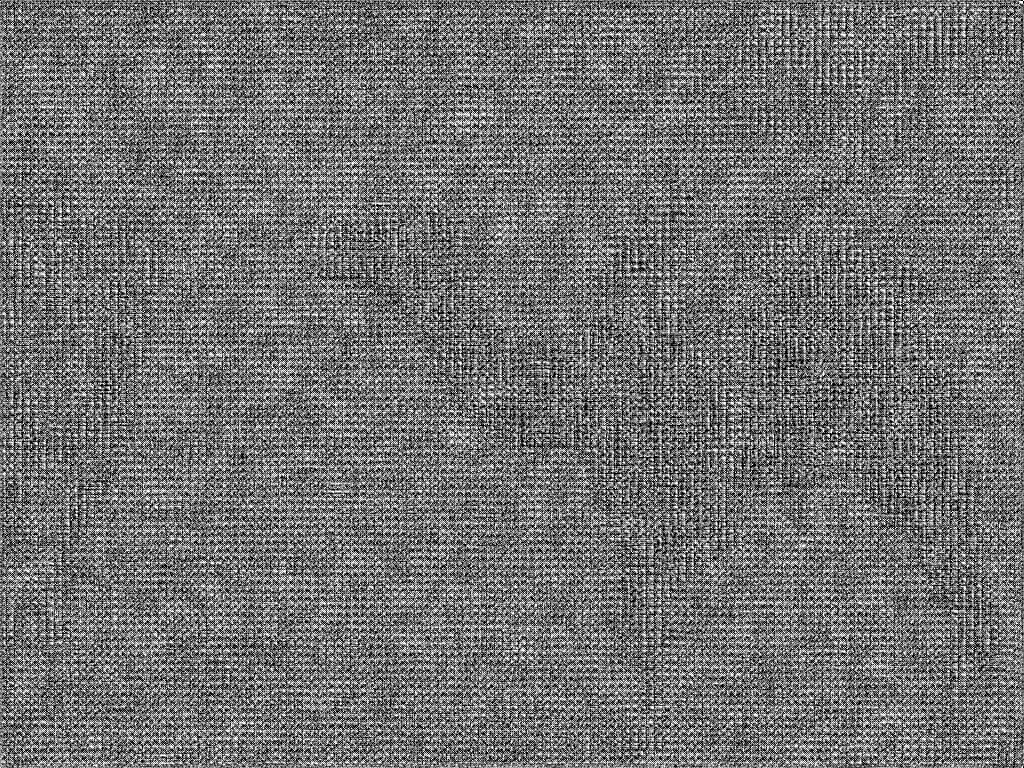}\hspace{2mm}
		\includegraphics[scale=0.11]{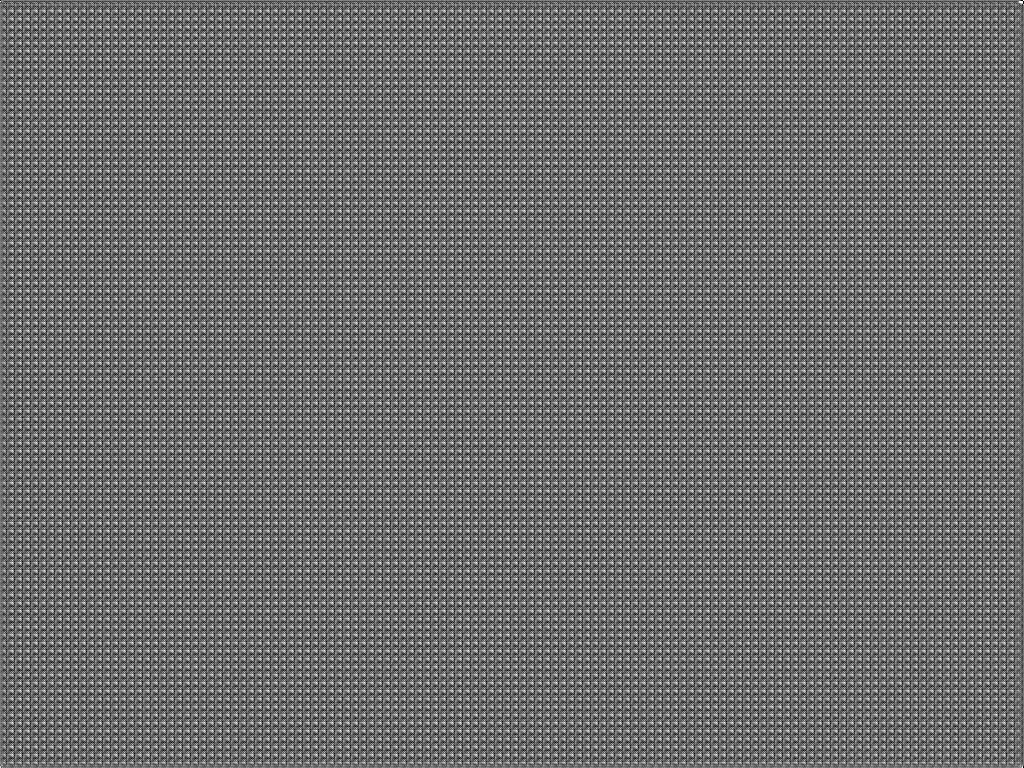}
	\end{tabular}
	\caption{Single-image (left) and 200-images (right) noiseprint estimates of the same camera.
    In the first case, some scene contents (camel) leaked in the estimate. In the second case, only periodic model-related artifacts are visible.}
	\label{fig:1}
\end{figure}

In the reminder of the paper we give some more details on noiseprint,
then describe the proposed noiseprint-based forgery localization procedure,
finally we discuss experimental results and draw conclusions.


\section{Noiseprint}

In \cite{Cozzolino_noiseprint} we set the goal of extracting a noise-like image, called noiseprint,
which works as a camera {\it model} fingerprint, much like the PRNU pattern can be regarded as a {\it device} fingerprint.
In principle, this could be obtained by simply keeping the noise residual without removing non-unique artifacts.
In practice, however, this residual does not possess the desired properties,
because it suffers from the very same problems as the PRNU, caused by significant leakages from the high-level signal.
Therefore, we designed a new system, based on deep learning, specifically dedicated to our goal.

To gain insight into our design choices, let us start from the ultimate goal.
Our system must accept in input a generic image and provide in output a residual image, the noiseprint,
having the same size as the input and containing only camera-model specific features with their natural spatial distribution.
This task reminds of residual-based denoisers,
which extract the additive Gaussian noise component of a given input image, thus removing the high-level content.
Such denoisers are trained by back-propagating the error between the the noise residual output by the CNN and the {\em true} noise pattern.
In our problem, however, no true reference is available for training, as the ideal noiseprint of a camera model is unknown.

Nonetheless, we know that images acquired by the same camera model have the same noiseprint.
We exploit this information in the training phase by
using {\em pairs} of images, extracting their residuals in parallel, and computing their distance (mean squared error).
When the pair comes from the same model, the error is back-propagated through the network to reduce the distance between residuals.
On the contrary, when images come from different models, back-propagation is used to increase the residual distance.
Therefore, it is like considering identical twin networks, working in parallel.
Each one uses the output of the other twin, together with the same/different label, to adapt its weights towards convergence.

The idea of Siamese networks is not new in deep learning.
Typically, each twin net extracts a low-dimensional vector (embedding) which summarizes the input keeping the relevant information for the specific task.
The main goal is to perform a low-complexity comparison of the twin outputs.
In our case, however, the aim is not reducing complexity, while it is important to preserve spatial information.
Hence, our net must preserve image size and spatial relationships.

\begin{figure}[t!]
	\centering
	\includegraphics[width=0.99\linewidth,page=1,trim={0 1cm 0 0}]{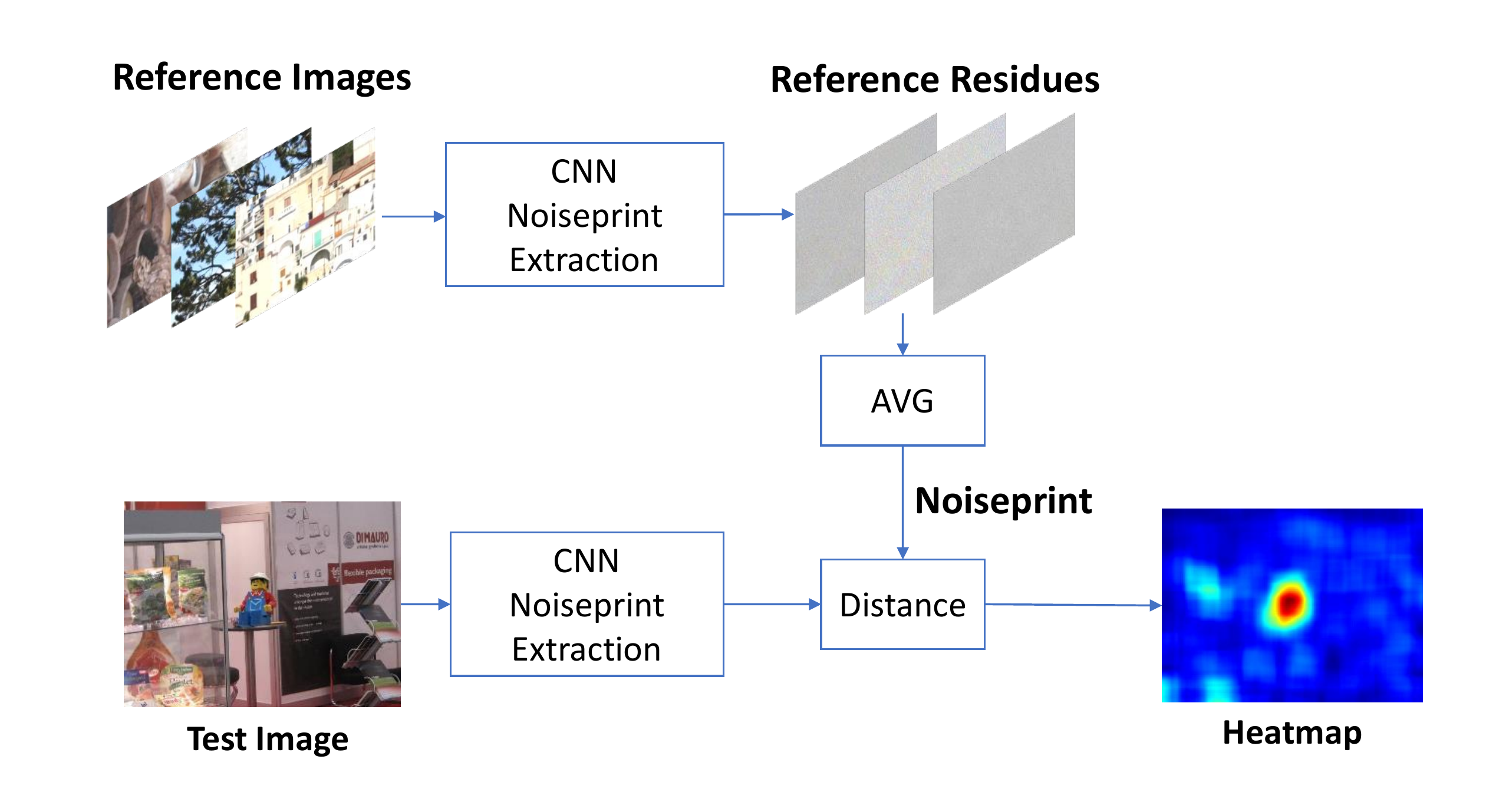}
	\caption{Proposed supervised forgery localization scheme.}
	\label{fig:Test}
\end{figure}

\begin{figure*}
	\centering
	\vspace{2mm}
	\begin{tabular}{cccc}
		\includegraphics[width = 0.22\linewidth]{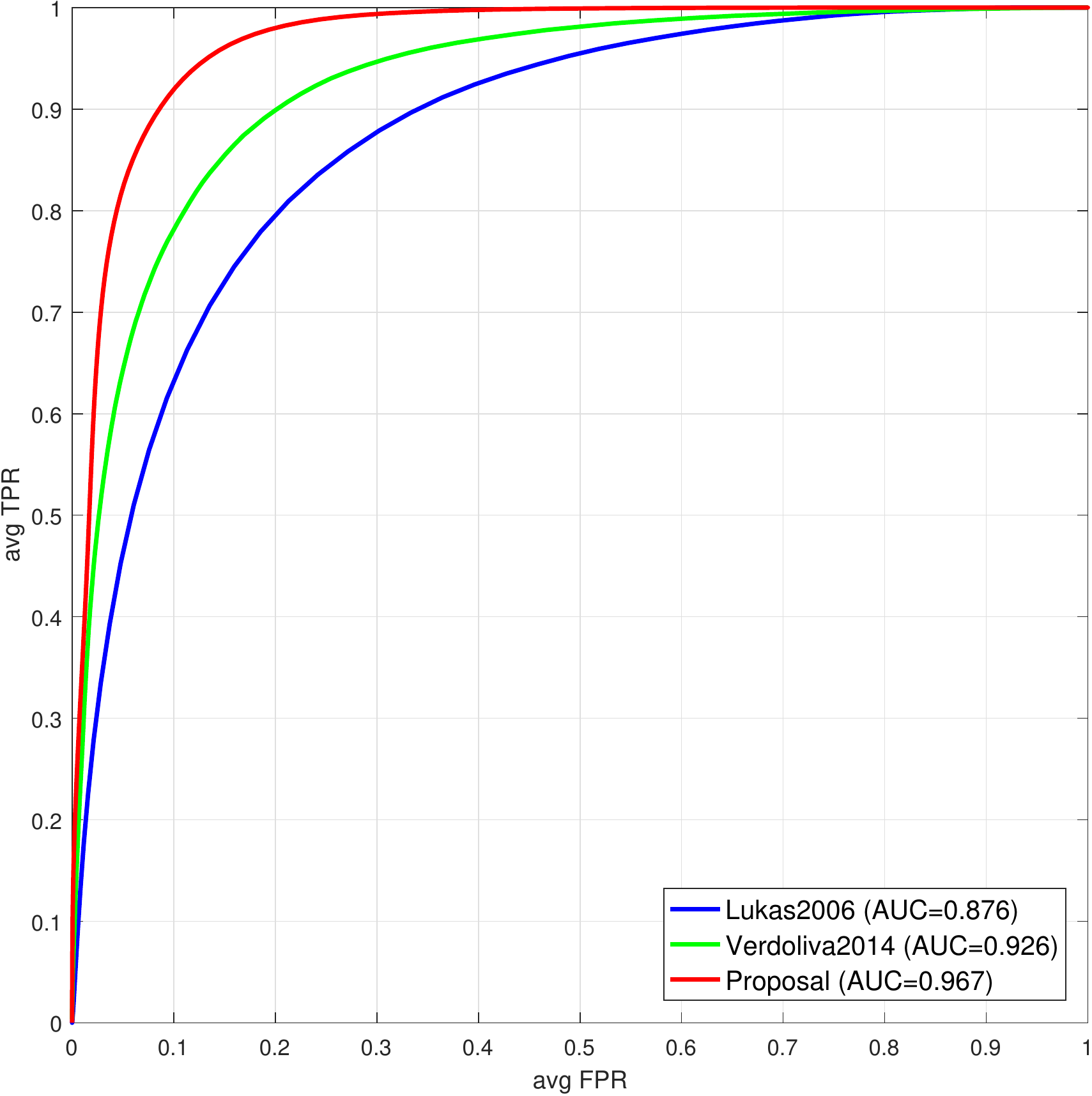} &
		\includegraphics[width = 0.22\linewidth]{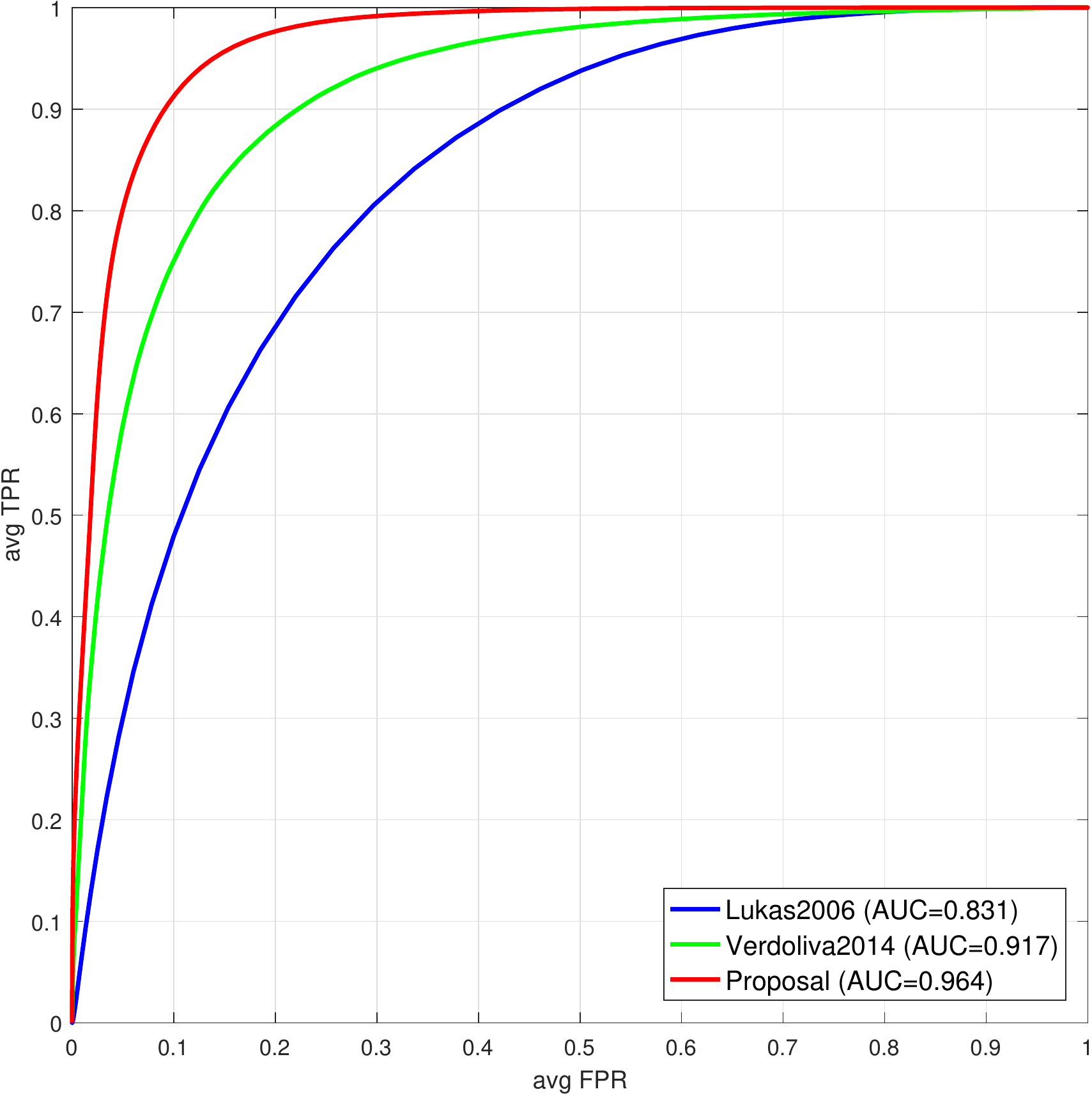} &
		\includegraphics[width = 0.22\linewidth]{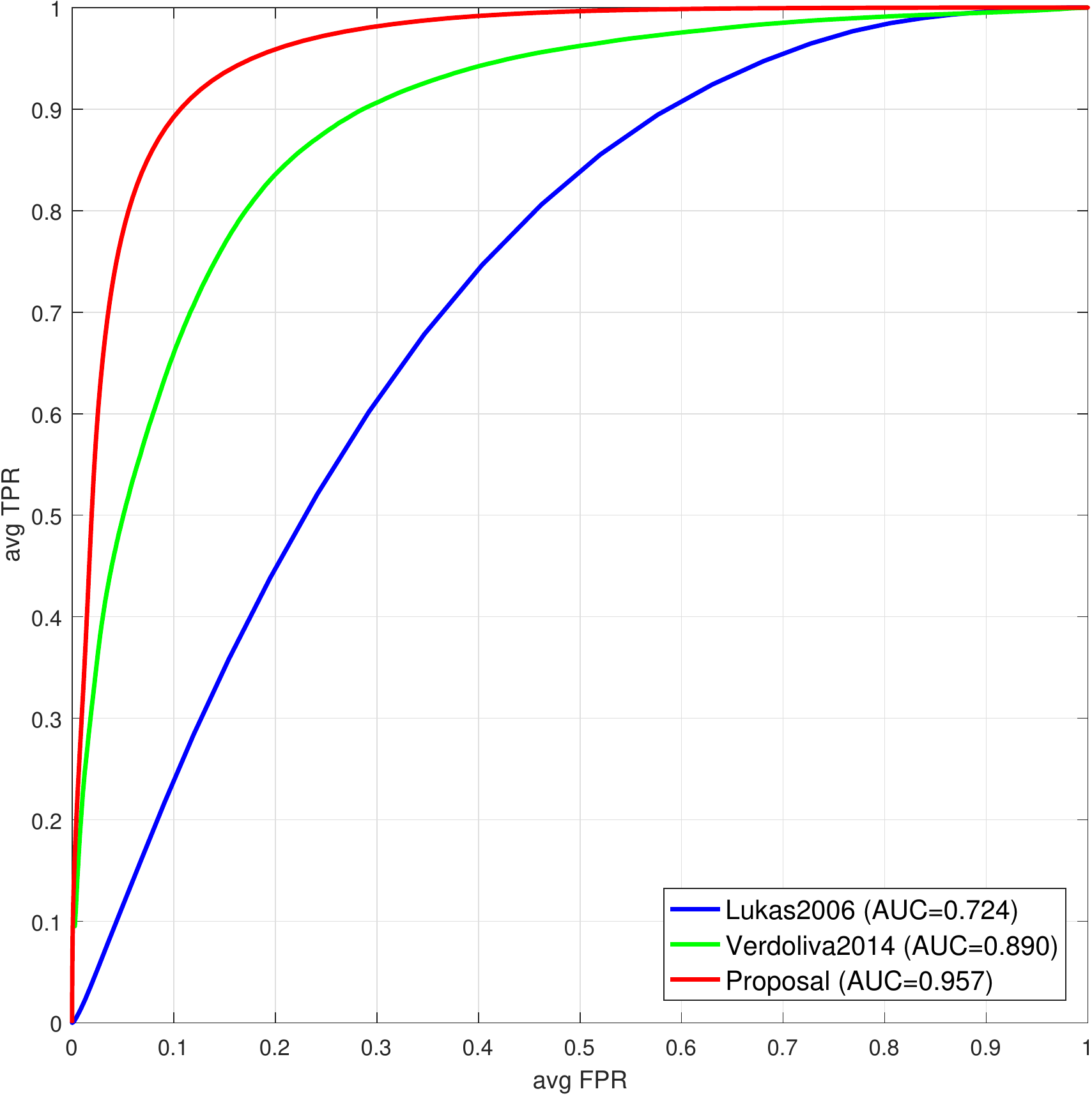} &
		\includegraphics[width = 0.22\linewidth]{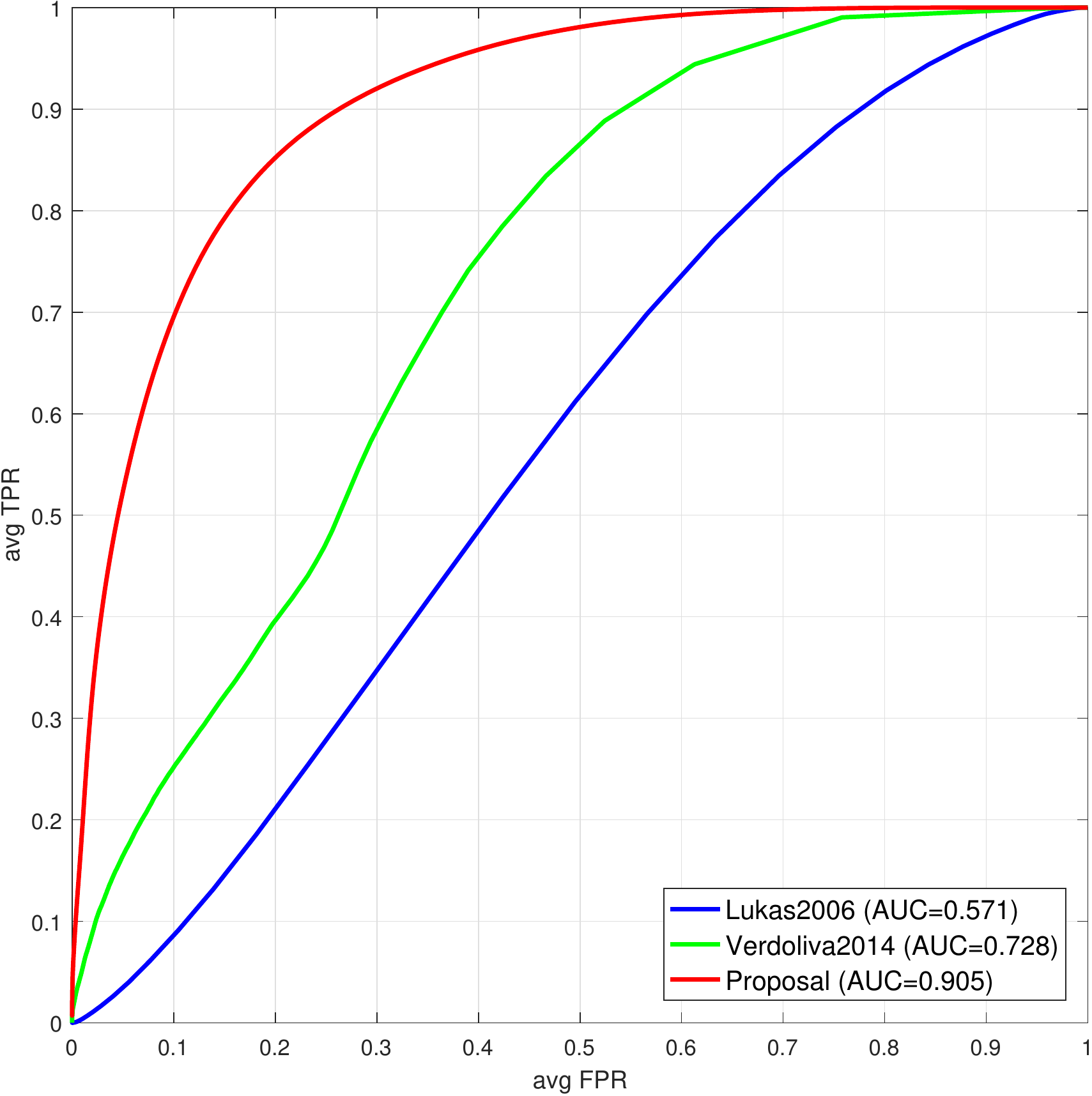} \\
	\end{tabular}
	
	\caption{ROCs at pixel level by varying the number of images for estimation (200,50,10,1).}
	\label{fig:ex_con}
\end{figure*}

Turning to the practical implementation,
we initialize the network (ideally, the Siamese nets) as the CNN denoiser proposed in \cite{Zhang2017} for AWGN (additive white Gaussian noise) image denoising.
Removing the scene content, in fact, is a reasonable starting point for our goal.
To limit training complexity, and also to ensure flexibility w.r.t. input size, we work on small image patches, rather than whole images.
Hence, we feed the network with a large number of paired patches,
with label $+1$ when they refer to the same camera model {\em and} the same spatial location, and label $-1$ otherwise.
The constraint on the spatial location is necessary to preserve the precious spatial information.
In fact, even a single-pixel shift impacts heavily on the local statistics of the residual.
The distance between the CNN outputs of the two paired patches is then computed.
In the loss function, pairs with negative label (different model and/or different position) are penalized.
Therefore, at convergence the net should extract the same residual for positive samples, and different residuals otherwise.

It is worth underlining that the network is trained once and for all, it can process any camera, both inside and outside the training set, and is not specific of a given experiment, or task.
Therefore, once the training is over, the noiseprint is deterministically related to the original image.

We refer the interested reader to \cite{Cozzolino_noiseprint} for details on the training phase.
Here, we mention only two key solutions adopted to improve the training efficiency,
the method proposed in \cite{Song2016} to obtain $O(n^2)$ rather than $O(n)$ samples for each $n$-patch minibatch,
and the distance based logistic (DBL) loss proposed in \cite{Vo2016}.

%

\section{Using noiseprint in a supervised setting}

In \cite{Cozzolino_noiseprint} we performed forgery localization with no side information, by looking for anomalies in the image residual.
Here, we consider a supervised setting, assuming a reference noiseprint is available.
The localization procedure is outlined in Fig.\ref{fig:Test} and follows the classic pipeline used with PRNU-based methods.
We rely on a set of pristine images taken by the same camera of the image under analysis.
Their residuals are averaged to obtain a clean reference,
namely, a reliable estimate of the camera noiseprint where high-level scene leakages, as well as traces of the PRNU, are mostly removed.
Fig.1 shows a single-image estimate, in which traces of the high-level content can be easily spotted (see the camel image in Fig.4), and a 200-image estimate which is virtually noise-free.
This reference is then compared in sliding window modality with the residual of the image under test, 
using the Euclidean distance as a measure of similarity.
As usual in these cases, the window size impacts on the trade-off between resolution and reliability.
Here, a 64$\times$64 window is used.

The pixel-wise distance field can then be shown as a heatmap,
which can be provided to the end user for visual inspection or subject to a suitable post-processing to extract a binary decision map.
Large distances (red in the heatmaps) suggest that the original noiseprint has been corrupted,
that is, deleted (inpainting), replaced with the noiseprint of another camera (splicing), or even of the same camera but after some geometrical distortion (resizing, rotation, even simple displacement).
Therefore, even a rigid copy-move can be discovered,
unless the displacement in both the vertical and horizontal directions is a multiple of the noiseprint fundamental period.
It is worth underlining, however,
that this approach makes sense only if the test image is aligned with the pristine reference images.
If the test image is geometrically distorted or subject to heavy compression,
also the references should undergo the same processing chain to deliver a correct reeference noiseprint.


\section{Experimental Results}

The network used to extract all noiseprints is trained on a large variety of models.
To this end, we created a large dataset, including both cameras and smartphones,
using various publicly available datasets, plus some other private cameras.
In detail, we used
44 cameras from the Dresden dataset \cite{Gloe2010},
32 from Socrates dataset \cite{Galdi2017},
32 from VISION \cite{Shullani2017},
17 from our private dataset,
totaling 125 individual cameras from 70 different models and 19 brands.
In the experiments, this dataset is split in training-set and validation-set comprising 100 and 25 cameras, respectively.
All images are originally in JPEG format with a quality factor in the range [96-99].
The network is initialized with the weights of the denoising network of \cite{Zhang2017}.
During training, each minibatch contains 200 patches of 48$\times$48 pixels extracted from 100 different images of 25 different cameras.
In each batch, there are 50 sets, each one formed by 4 patches with same camera and position.
Training is performed using ADAM optimizer; the hyper-parameters
(learning rate, number of iterations and weight of regularization term) are chosen using the validation set.

For image forgery localization, we compare results with
the PRNU-based method proposed in \cite{Lukas2006} (Lukas2006) and with a feature-based approach relying on camera model artifacts \cite{Verdoliva2014} (Verdoliva2014).
We use the same 300-image dataset with splicings already used in \cite{Chierchia2014}.
Images come from 4 camera models (CanonEOS 450D, Nikon D200, CanonIXUS\_95IS, NikonCoolpix\_S5100) none of which used for training the network.

In Fig.3(left) we show pixel-level localization results in terms of receiver operating curves (ROC)
when 200 images are used to estimate the reference noiseprint, PRNU, or the statistic of \cite{Verdoliva2014}.
The proposed method provides a large gain over the Lukas2006, and also over Verdoliva2014.
Synthetic results reported in Tab.I, in terms of area under curve (AUC) and F-measure, fully confirm this analysis.
For F-measure we used both the best threshold over all the dataset and (more favourable for performance) the best threshold for each image (F1-oracle).
In all cases, the performance gain of the proposed method is clear.
The very same conclusions can be drawn by visual inspection of the examples shown in Fig.4,
where very different types of manipulation have been considered (splicing, rigid copy-move, inpainting).
Note that all images have the native camera JPEG quality, and no further compression is carried out.
The noiseprint-based method exhibits always a very good performance, highlighting clearly the manipulations without false alarms.

\begin{table}
	\caption{Pixel-level localization performance.}
	\centering
	\begin{tabular}{|l||c|c|c|}
		\hline
		\ru            & Lukas2006 & Verdoliva2014 & Proposed \\ \hline\hline
		\ru AUC        &     0.876 &         0.926 &    0.967 \\
		\ru F1         &     0.499 &         0.580 &    0.724 \\
		\ru F1-oracle  &     0.572 &         0.707 &    0.850 \\ \hline
	\end{tabular}
	\label{tab:MCC_cmp}
\end{table}

As we remarked several times, the noiseprint appears to be less noisy than the PRNU pattern.
Therefore, unlike for PRNU-based methdos, we may expect the performance to depend only weakly on the number of reference images.
To investigate this point, we repeated the previous experiment using only 50, 10, and 1 reference images, respectively.
The corresponding ROCs, also shown in Fig.3, confirm noiseprint robustness.
While the performance of the other methods are largely impaired when less than 50 images are used,
the proposed method works reasonably well even with a single reference image, and better than Lukas2006 with 200 images.


In the last experiment we consider a more challenging situation, 
where the images have a different format, and hence are not well aligned with our dataset.
We take some examples from the dataset used by Korus et al. \cite{Korus2017}.
The images are taken from 4 different camera models (Canon 60D, Sony A57, Nikon D7000 and Nikon D90) and attacked with different forms of manipulations.
They are in raw format, hence no JPEG artifacts are present, a significant mis-alignment with respect to the trained CNN.
We use the PRNU provided with this dataset, estimated over 200 natural images for Nikon and Canon, and over 20 flat images for Sony.
For noiseprint and for \cite{Verdoliva2014} we used only the 53 available pristine images,
taking care to avoid using the same background in training and test.
As expected, Lukas2006 works better than in the previous case, since images are not compressed,
while the performance of the proposed method is impaired due to the mis-alignment.
Even in this critical situation, however, the noiseprint-based method keeps providing valuable hints for forgery localization.

\begin{figure*}
	\centering
	\begin{tabular}{c@{\hspace{2mm}}c@{\hspace{2mm}}c@{\hspace{2mm}}c}    \rule{0mm}{00mm}
		\includegraphics[width = 0.20\linewidth]{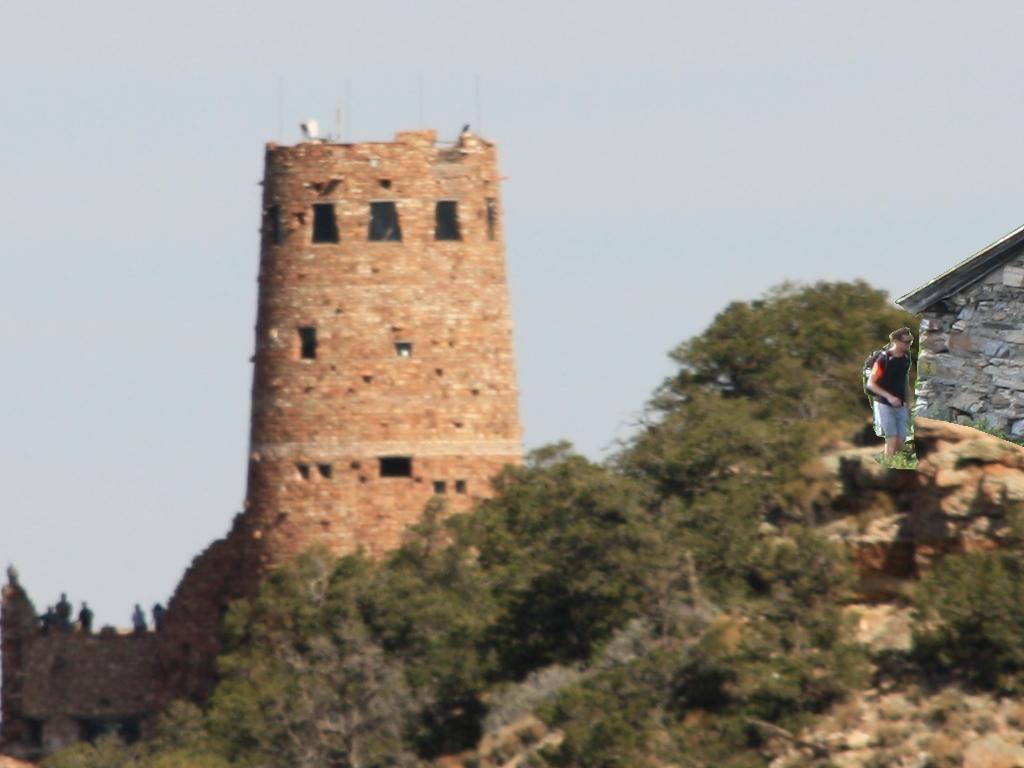}  &
		\includegraphics[width = 0.20\linewidth]{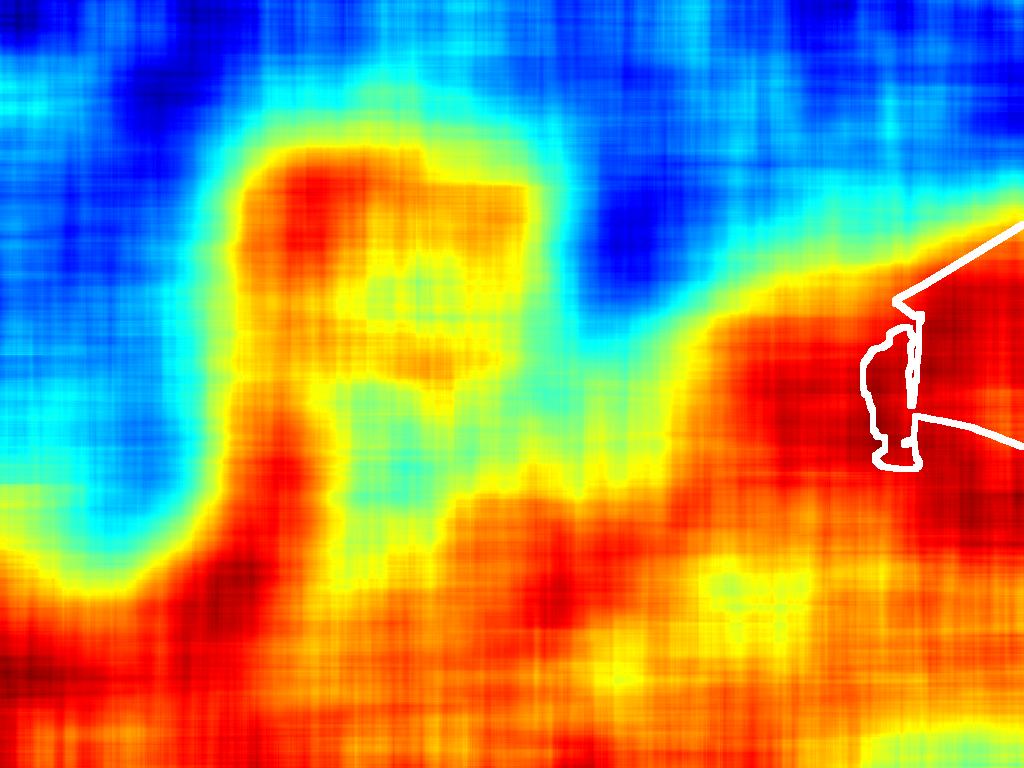} &
		\includegraphics[width = 0.20\linewidth]{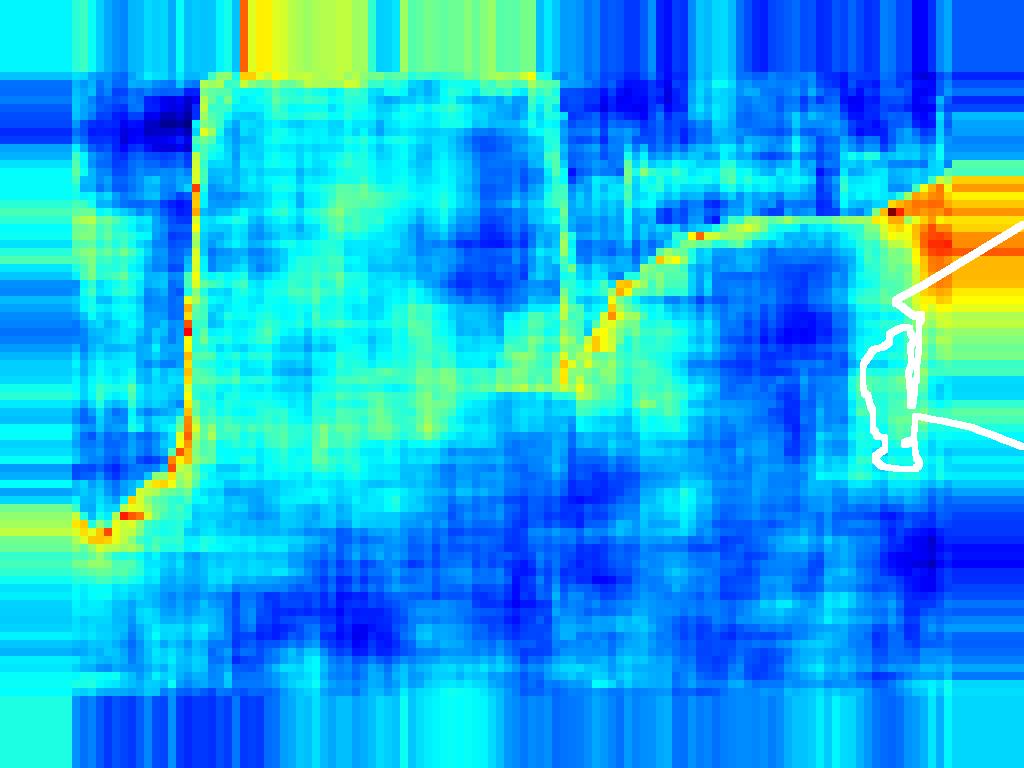} &
		\includegraphics[width = 0.20\linewidth]{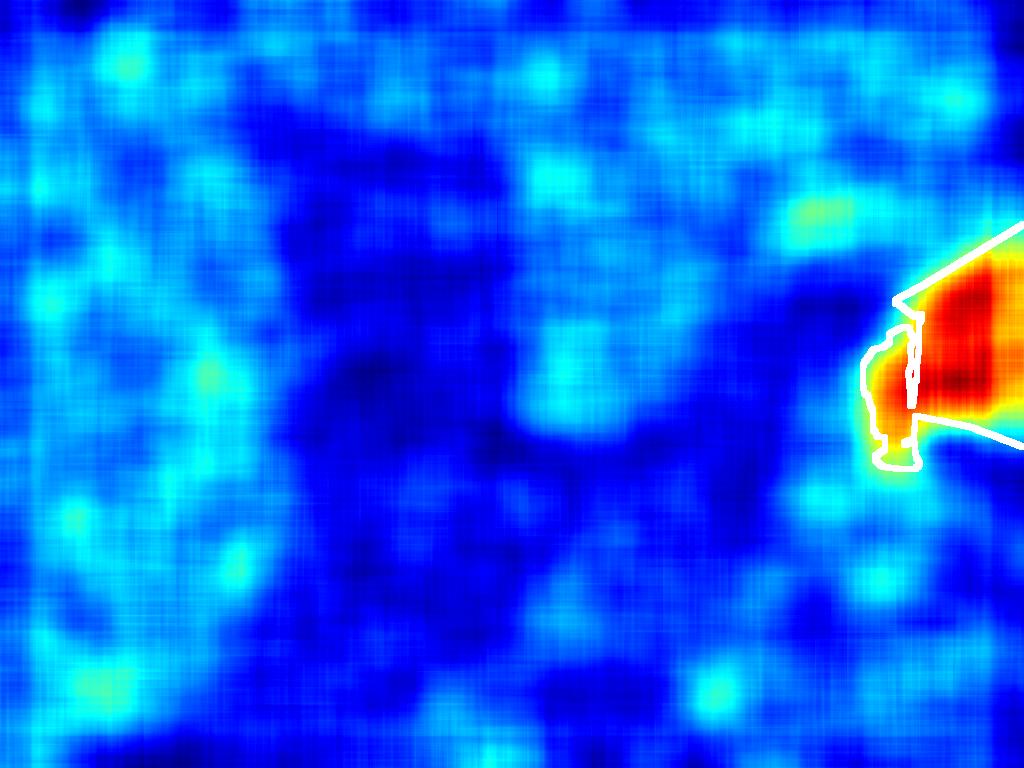} \\ \rule{0mm}{28mm}
		\includegraphics[width = 0.20\linewidth]{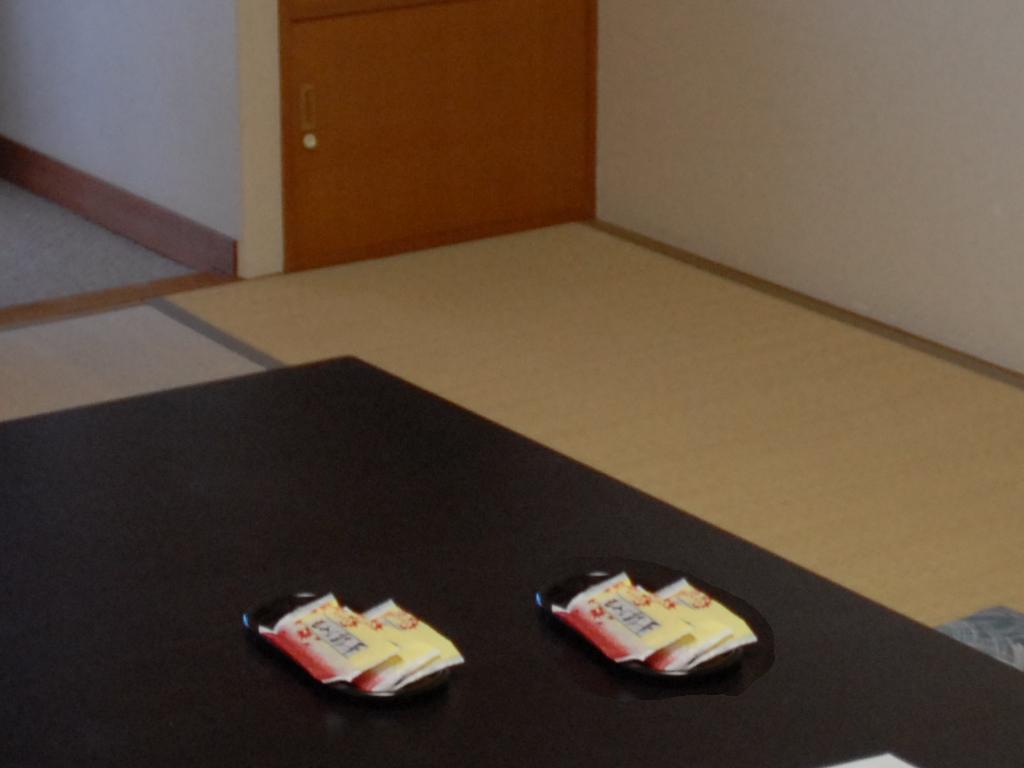}  &
		\includegraphics[width = 0.20\linewidth]{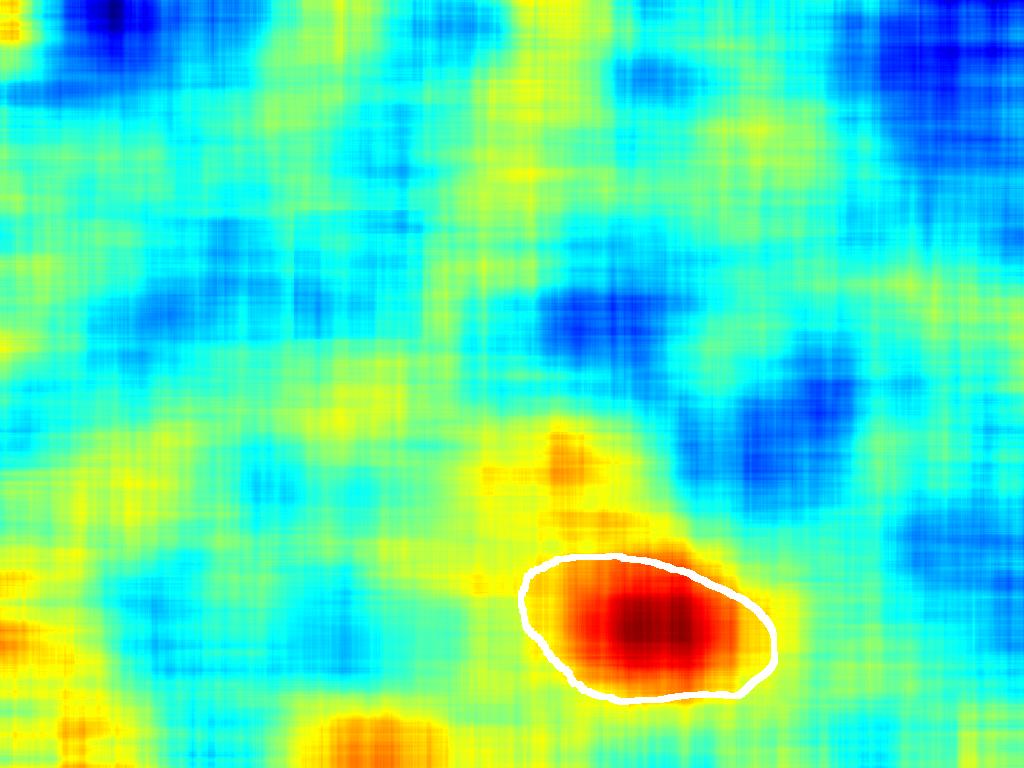} &
		\includegraphics[width = 0.20\linewidth]{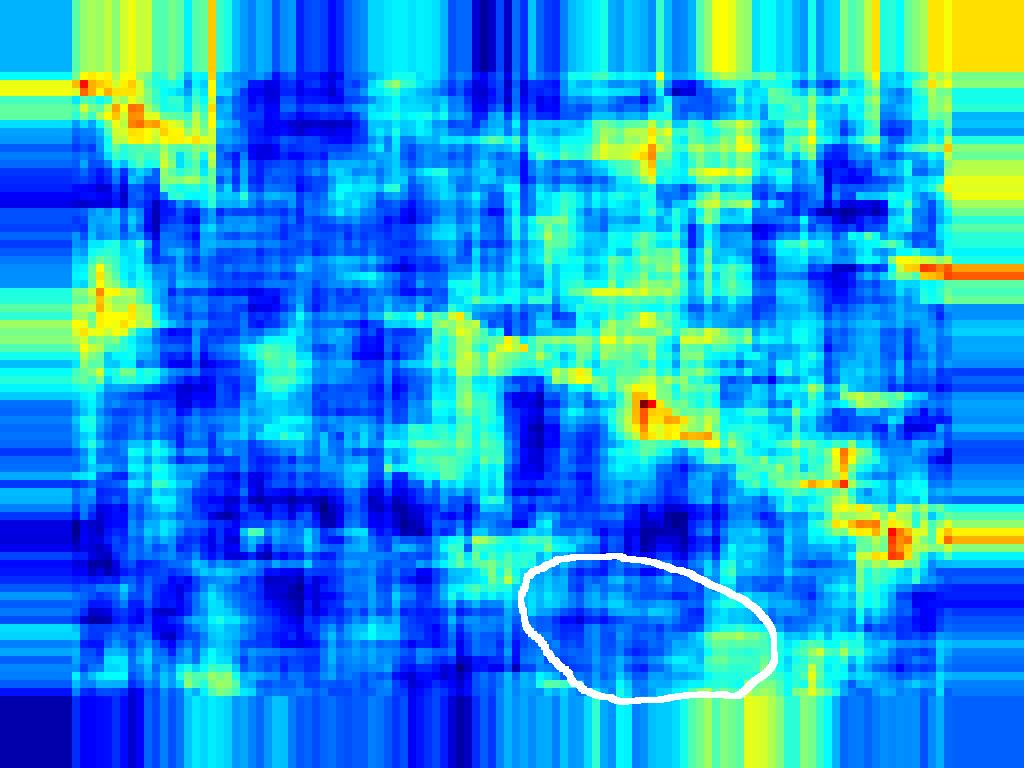} &
		\includegraphics[width = 0.20\linewidth]{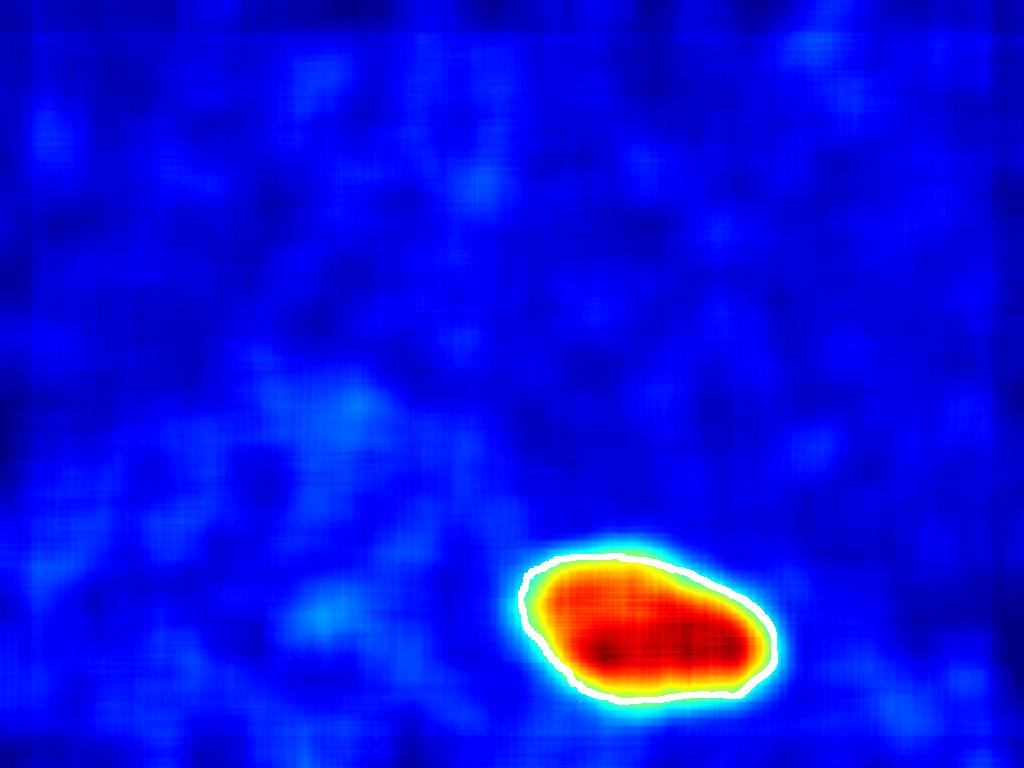} \\ \rule{0mm}{28mm}
		\includegraphics[width = 0.20\linewidth]{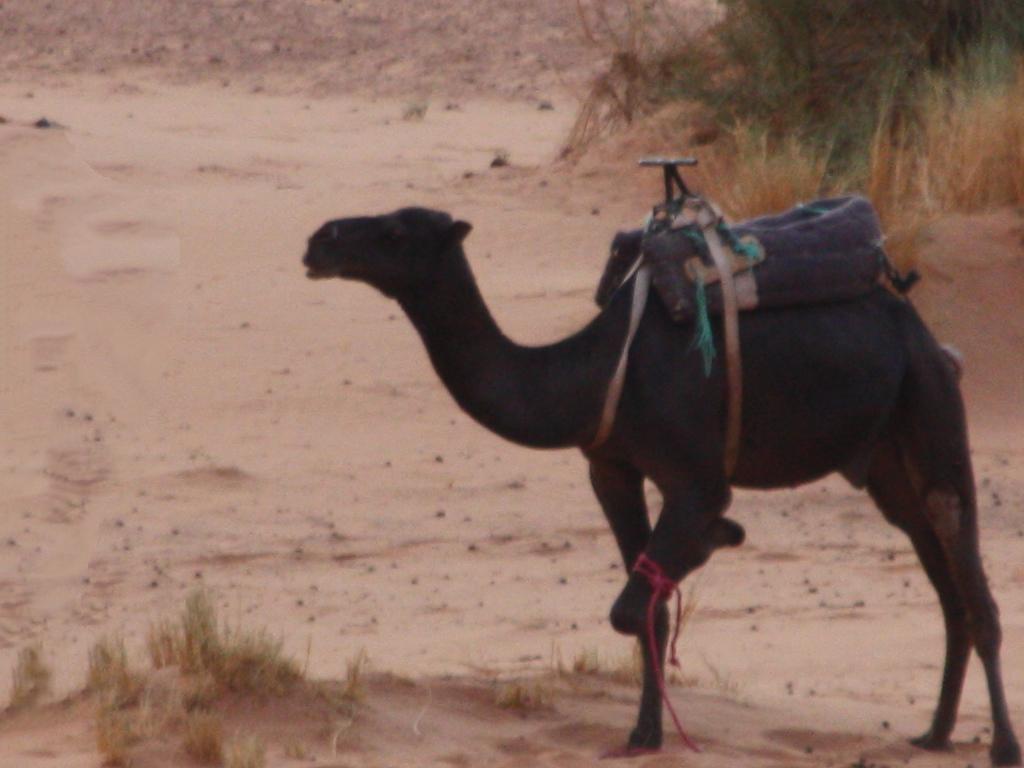}  &
		\includegraphics[width = 0.20\linewidth]{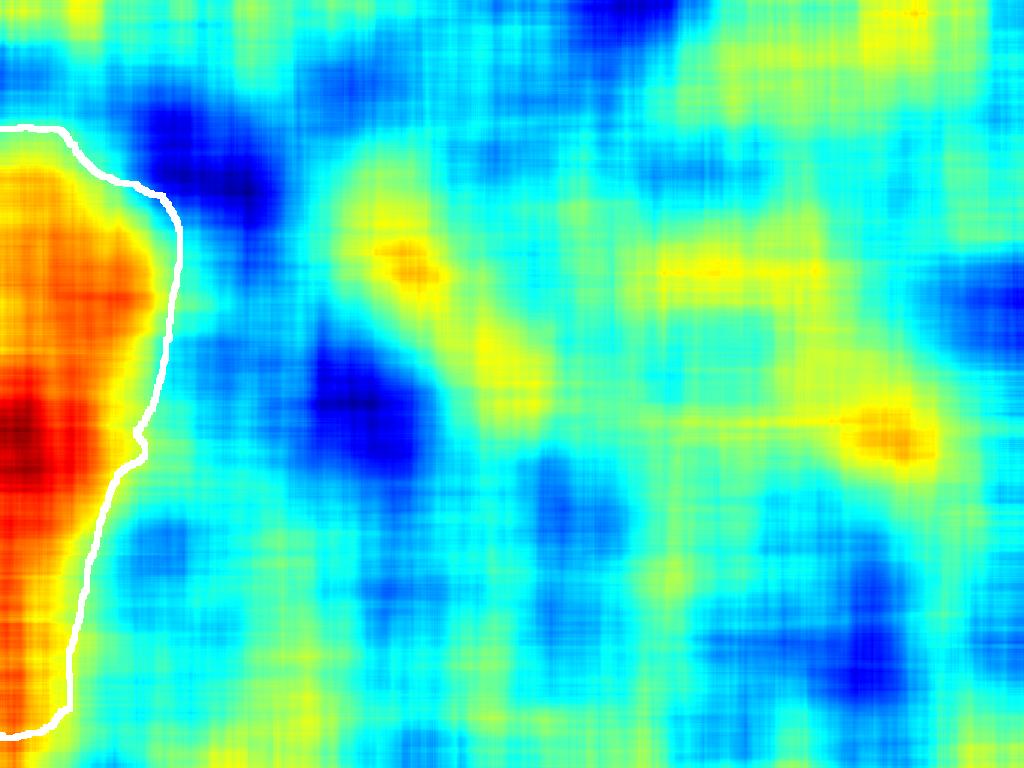} &
		\includegraphics[width = 0.20\linewidth]{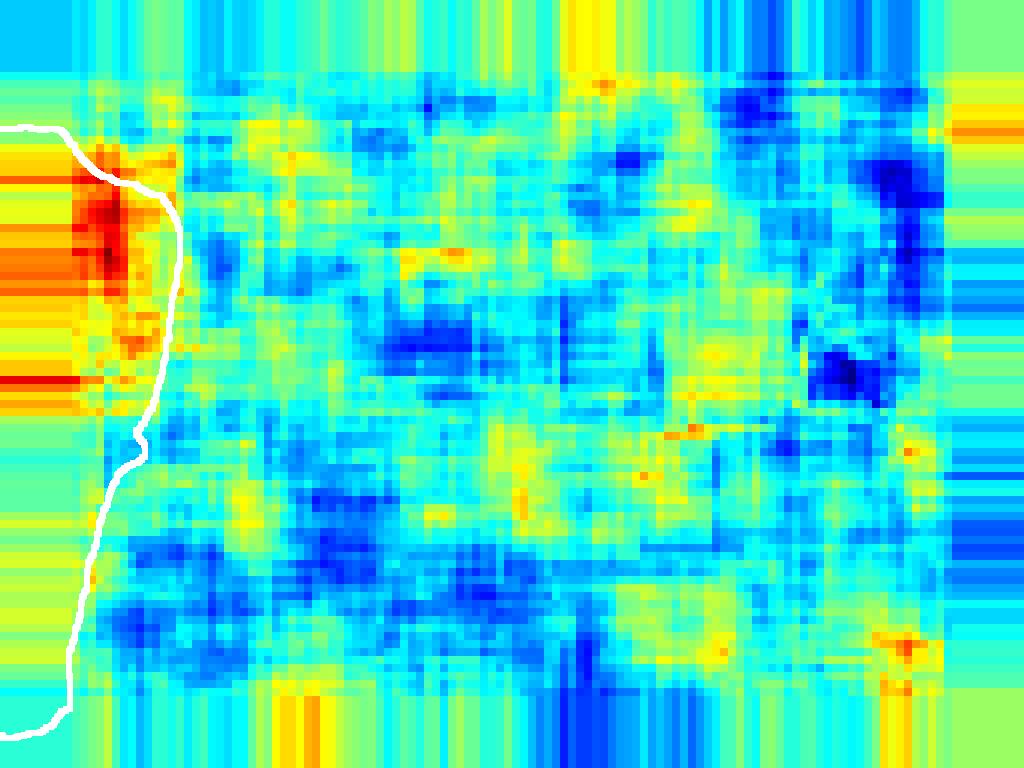} &
		\includegraphics[width = 0.20\linewidth]{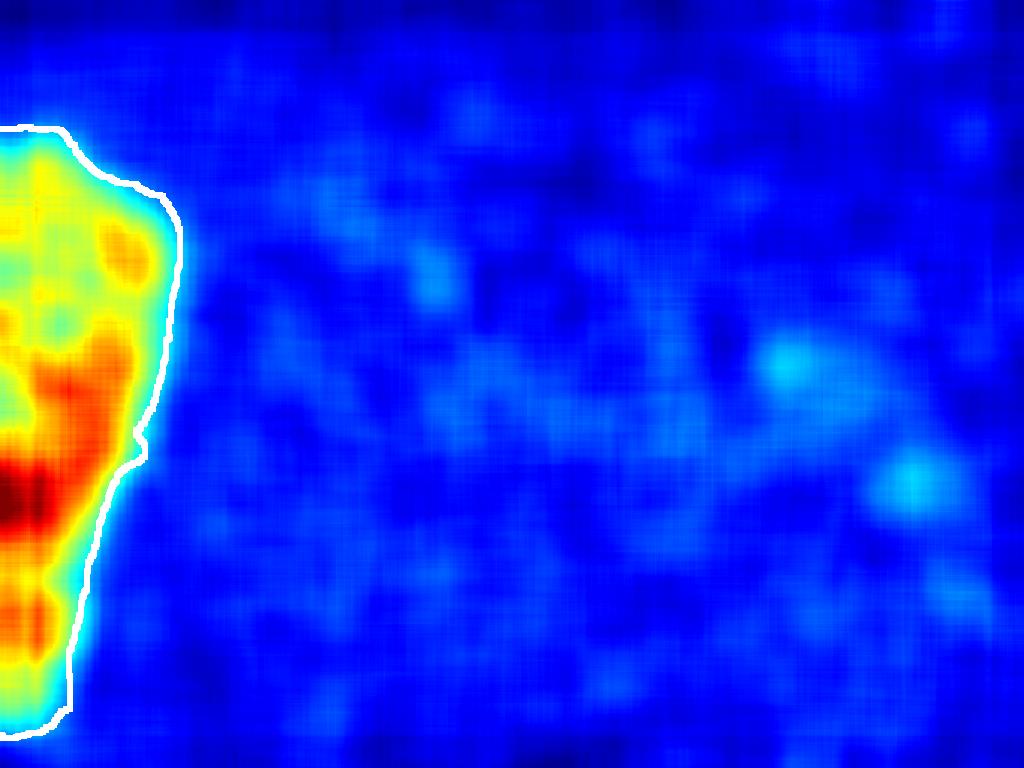} \\ 
		{\small (a) Forged image} &
		{\small (b) Lukas2006} &
		{\small (c) Verdoliva2014} &
		{\small (d) Proposed} \\
	\end{tabular}
	\caption{Forgery localization results for some selected examples. From top to bottom: splicing, rigid copy-move, inpainting.}
	\label{fig:ex_con}
\end{figure*}

\begin{figure*}
	\centering
	\begin{tabular}{c@{\hspace{2mm}}c@{\hspace{2mm}}c@{\hspace{2mm}}c}          \rule{0mm}{00mm}
		\includegraphics[width = 0.23\linewidth]{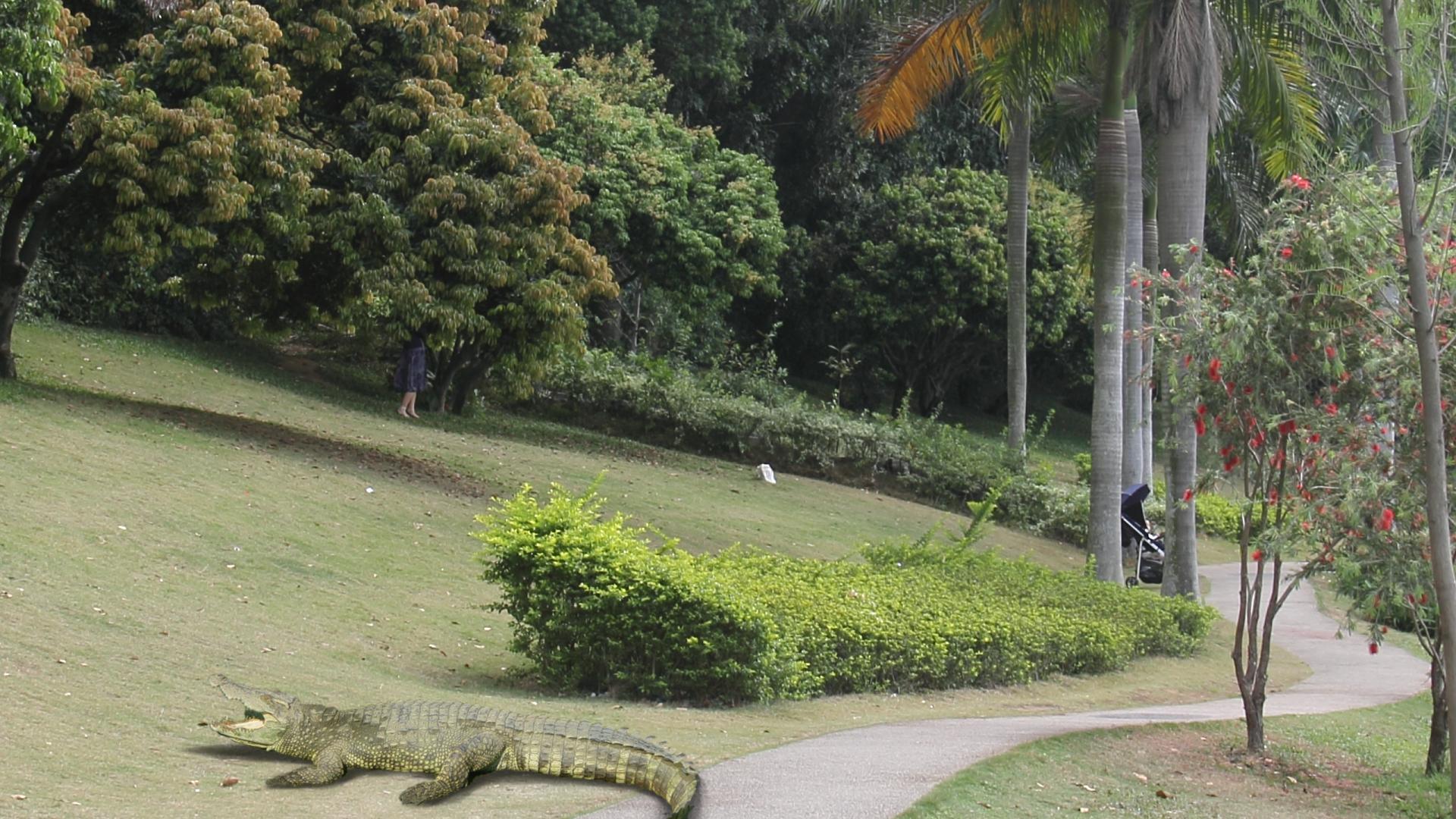}  &
		\includegraphics[width = 0.23\linewidth]{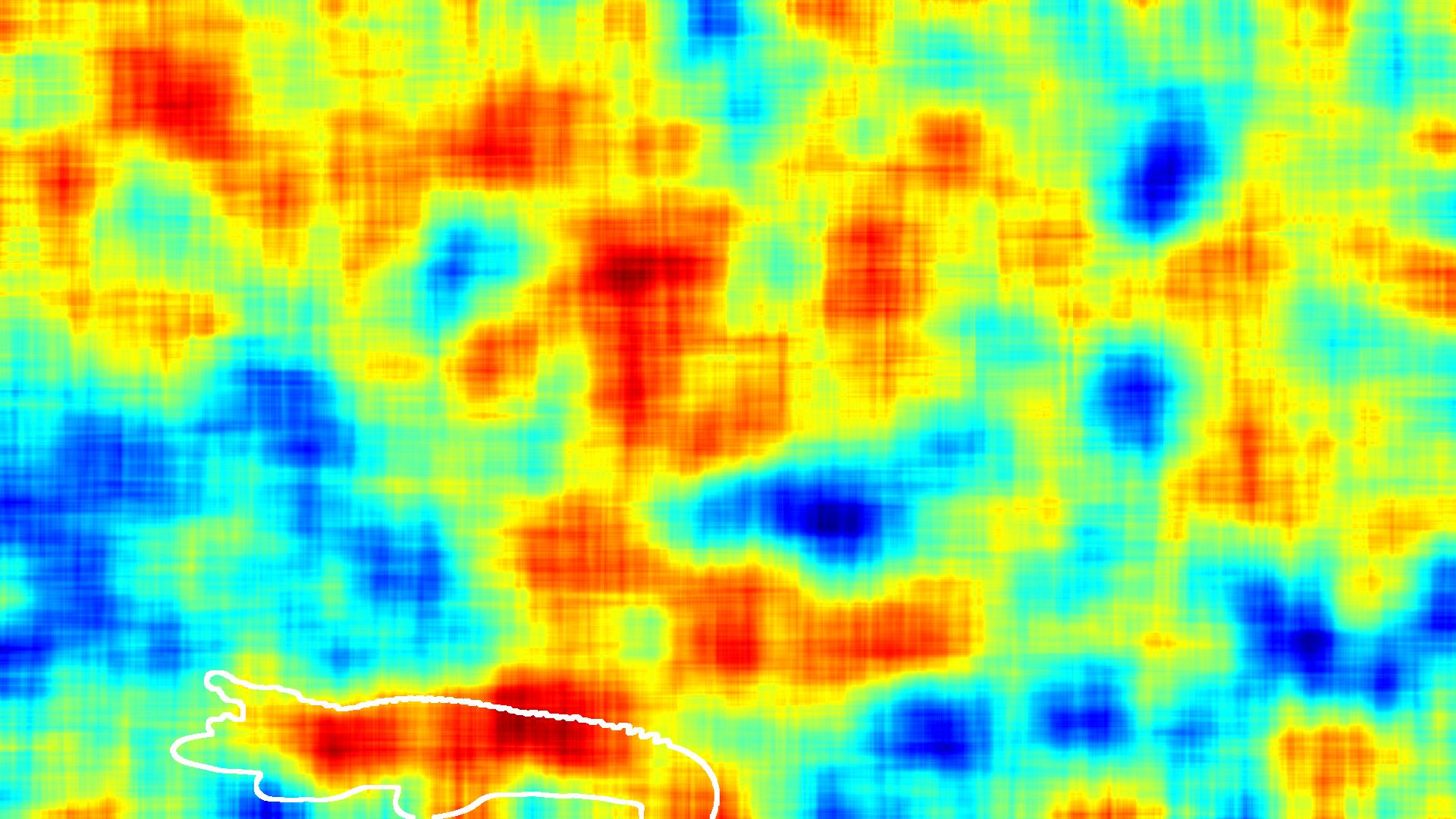} &
		\includegraphics[width = 0.23\linewidth]{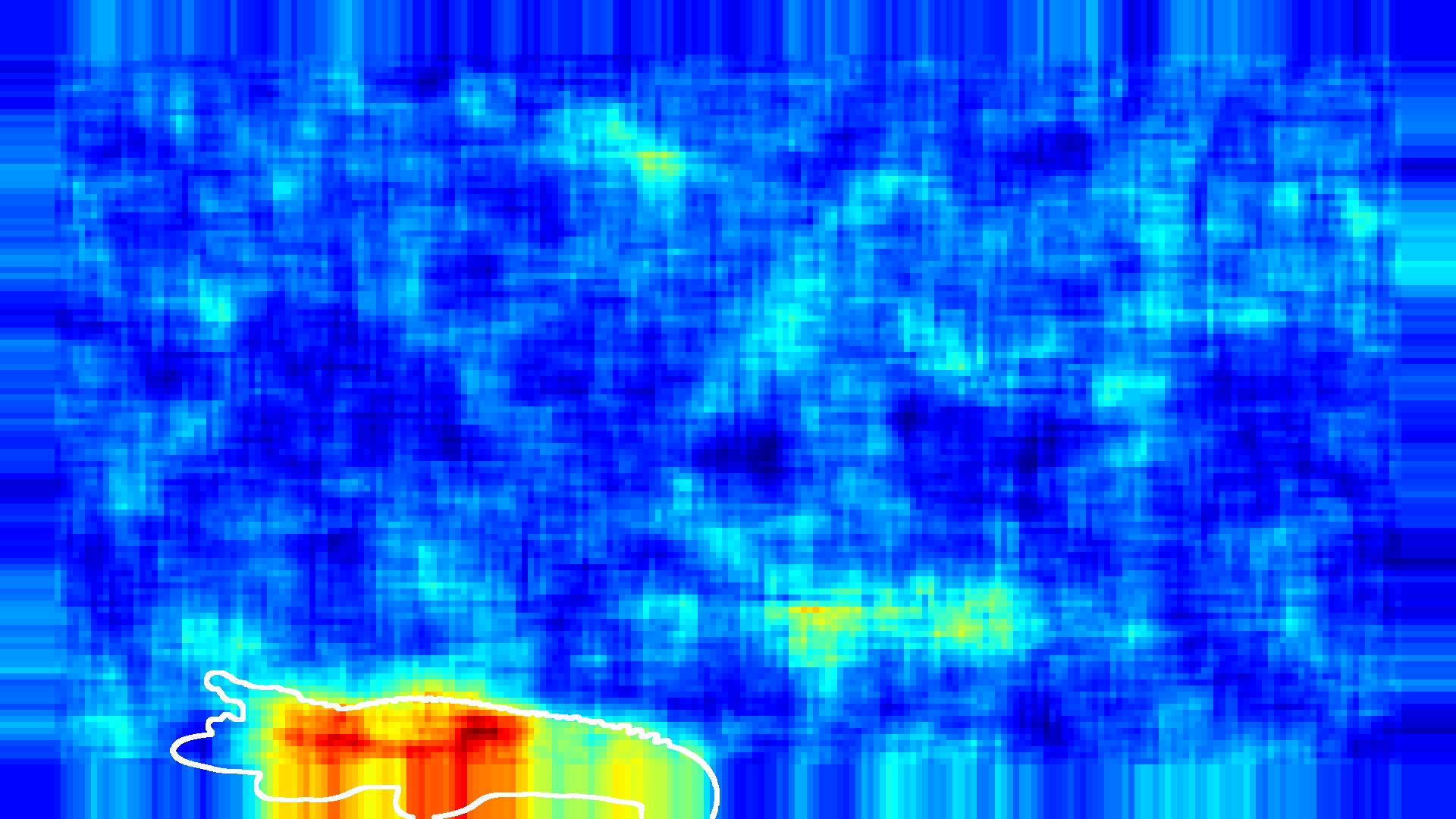} &
		\includegraphics[width = 0.23\linewidth]{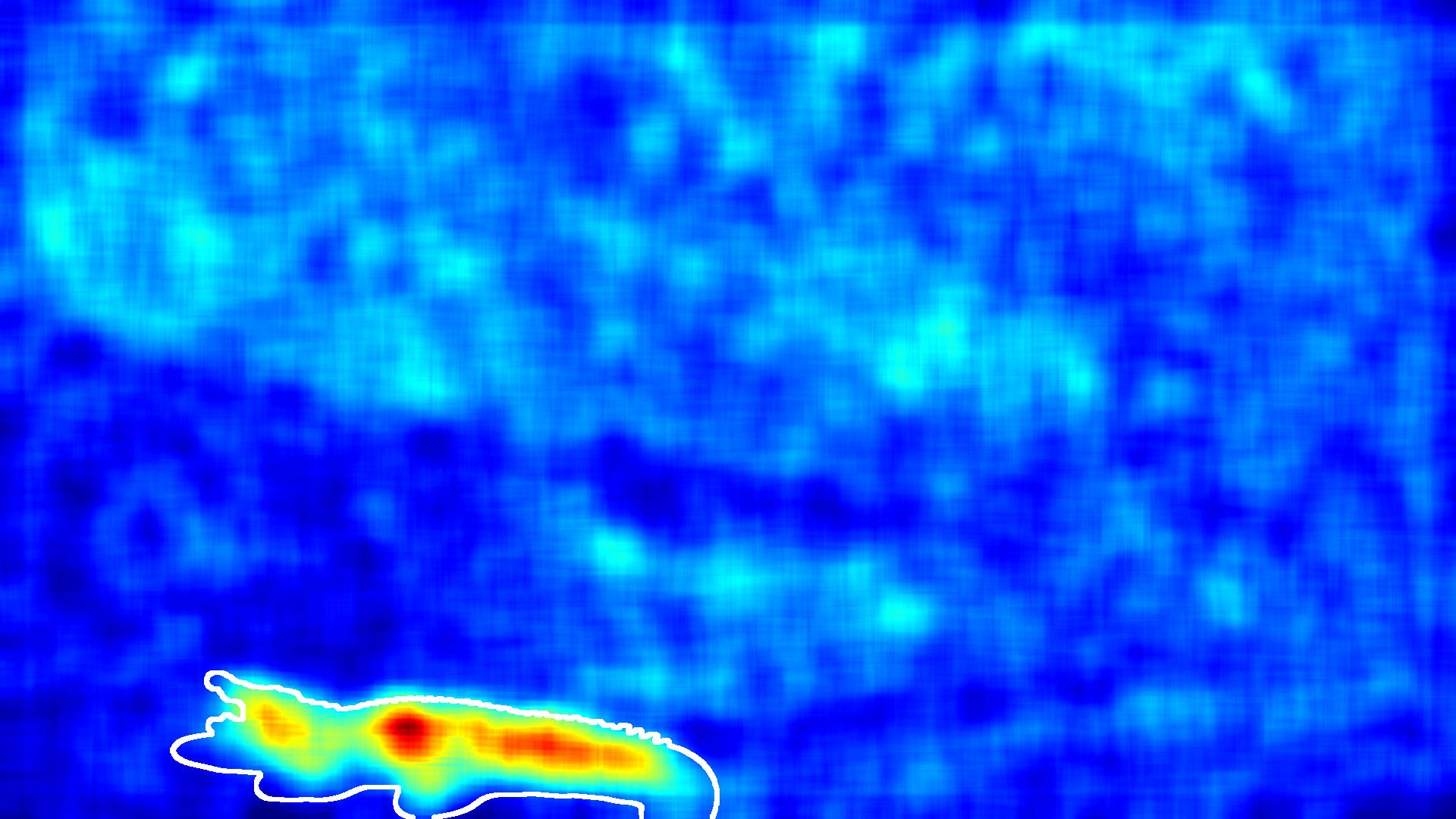} \\ \rule{0mm}{24mm}
		\includegraphics[width = 0.23\linewidth]{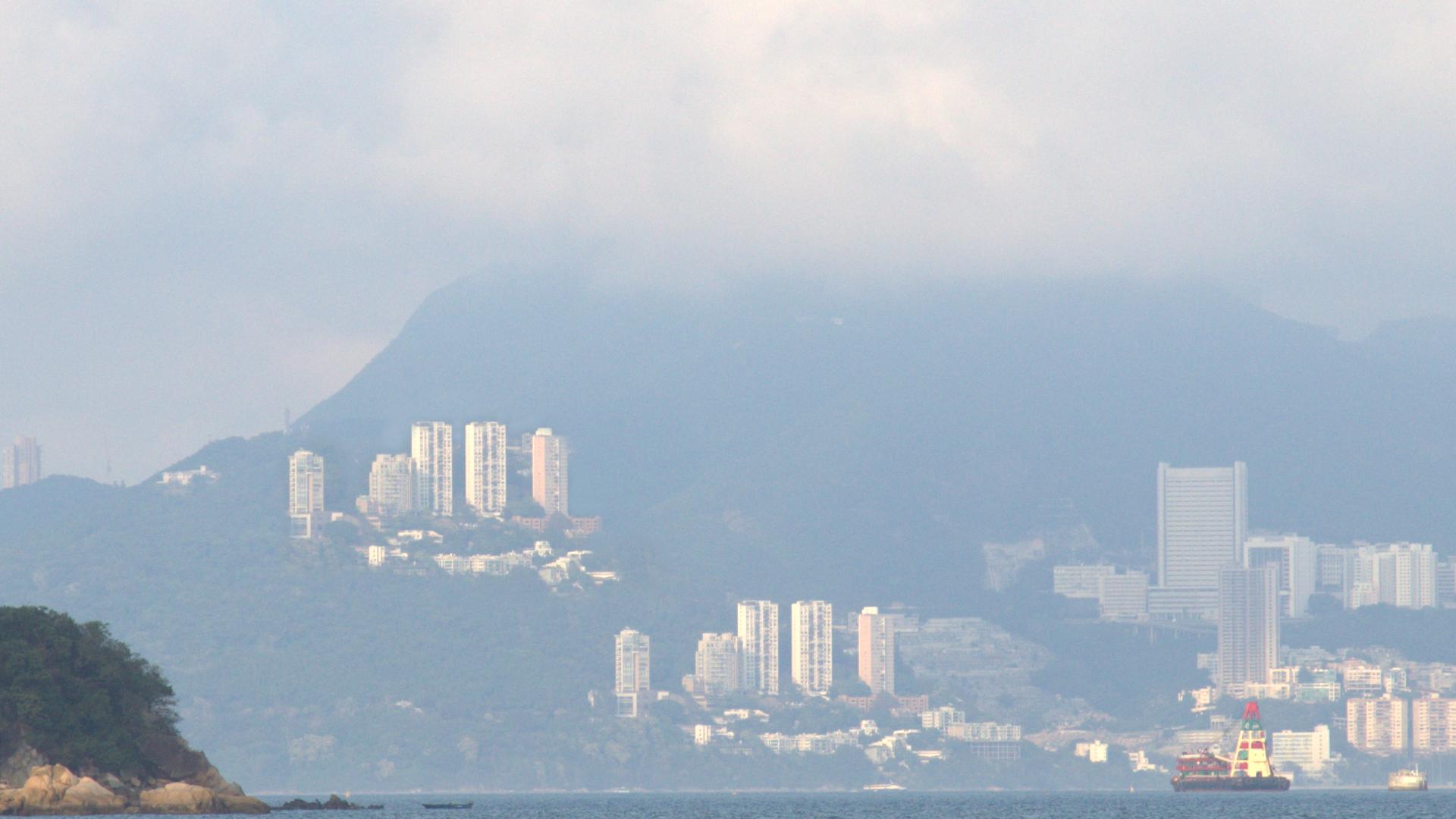}  &
		\includegraphics[width = 0.23\linewidth]{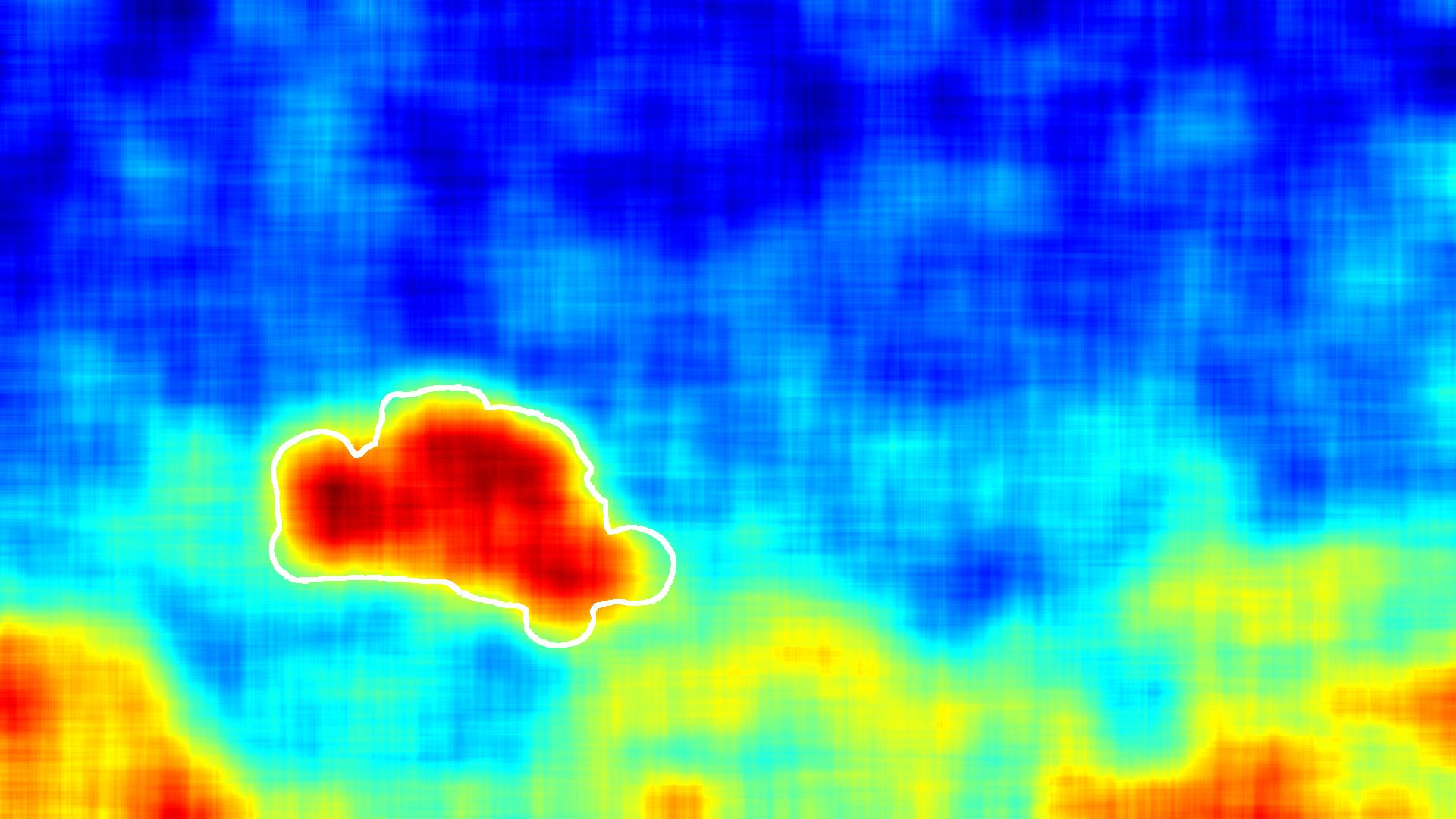} &
		\includegraphics[width = 0.23\linewidth]{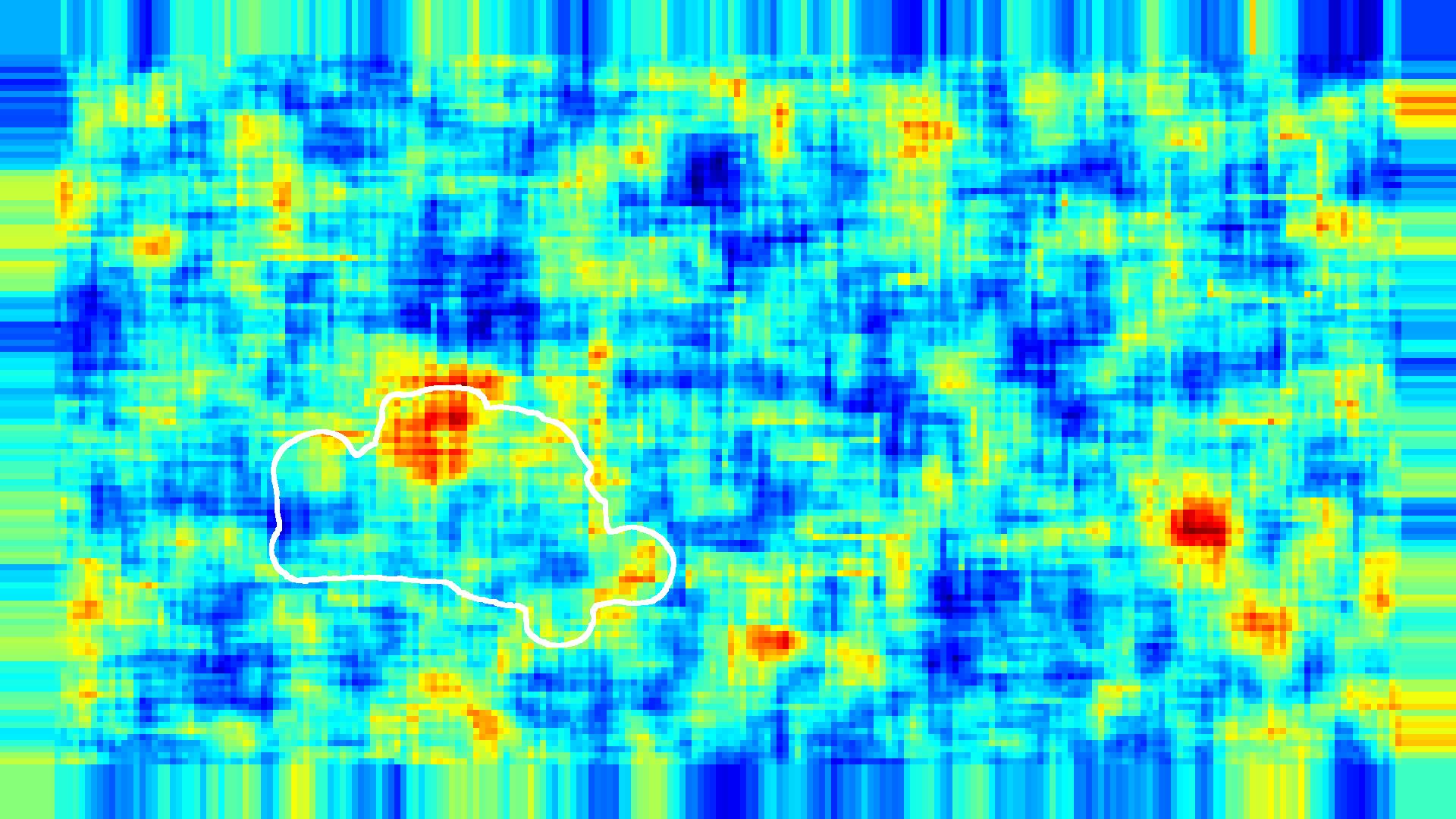} &
		\includegraphics[width = 0.23\linewidth]{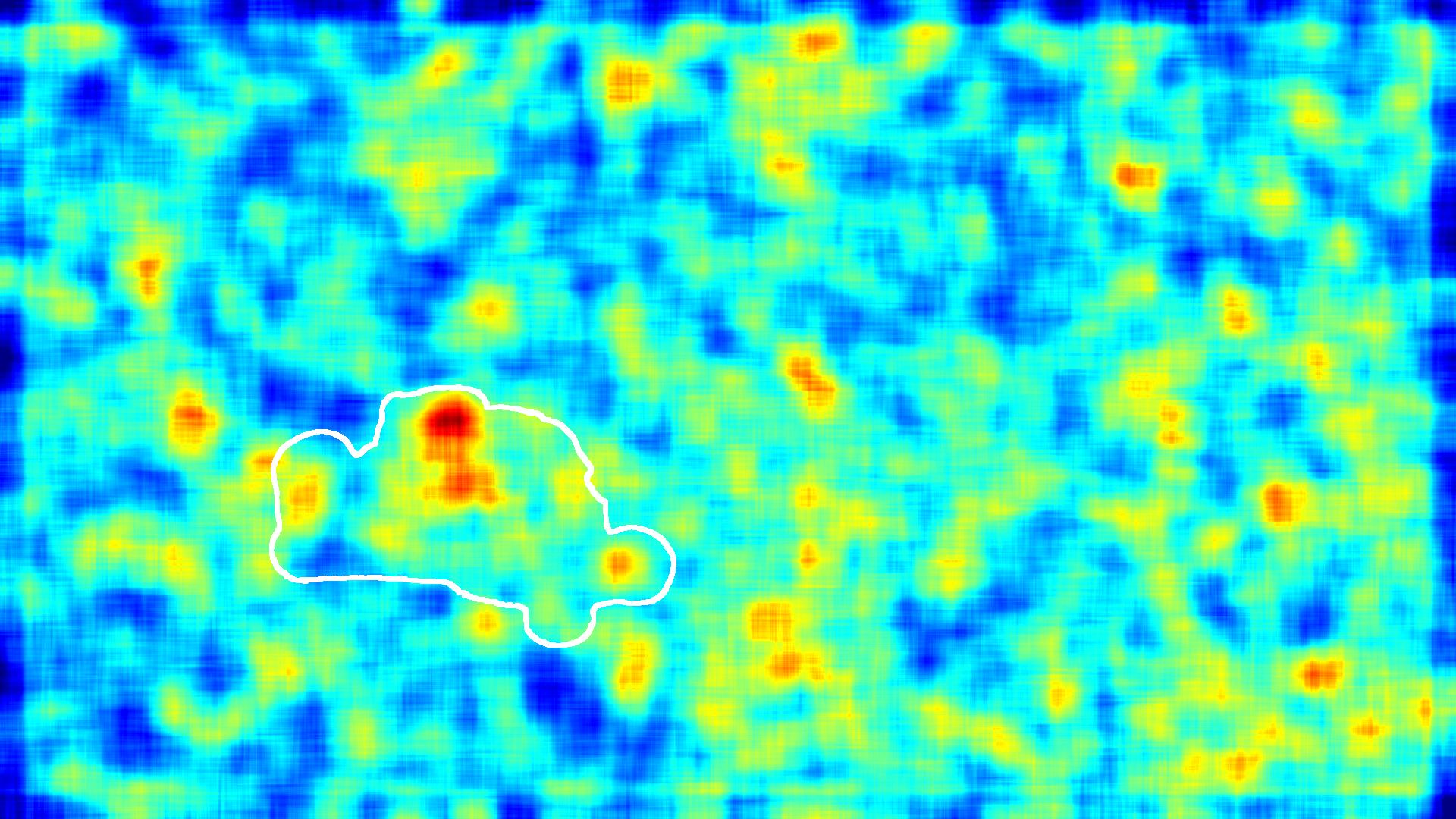} \\ \rule{0mm}{24mm}
		\includegraphics[width = 0.23\linewidth]{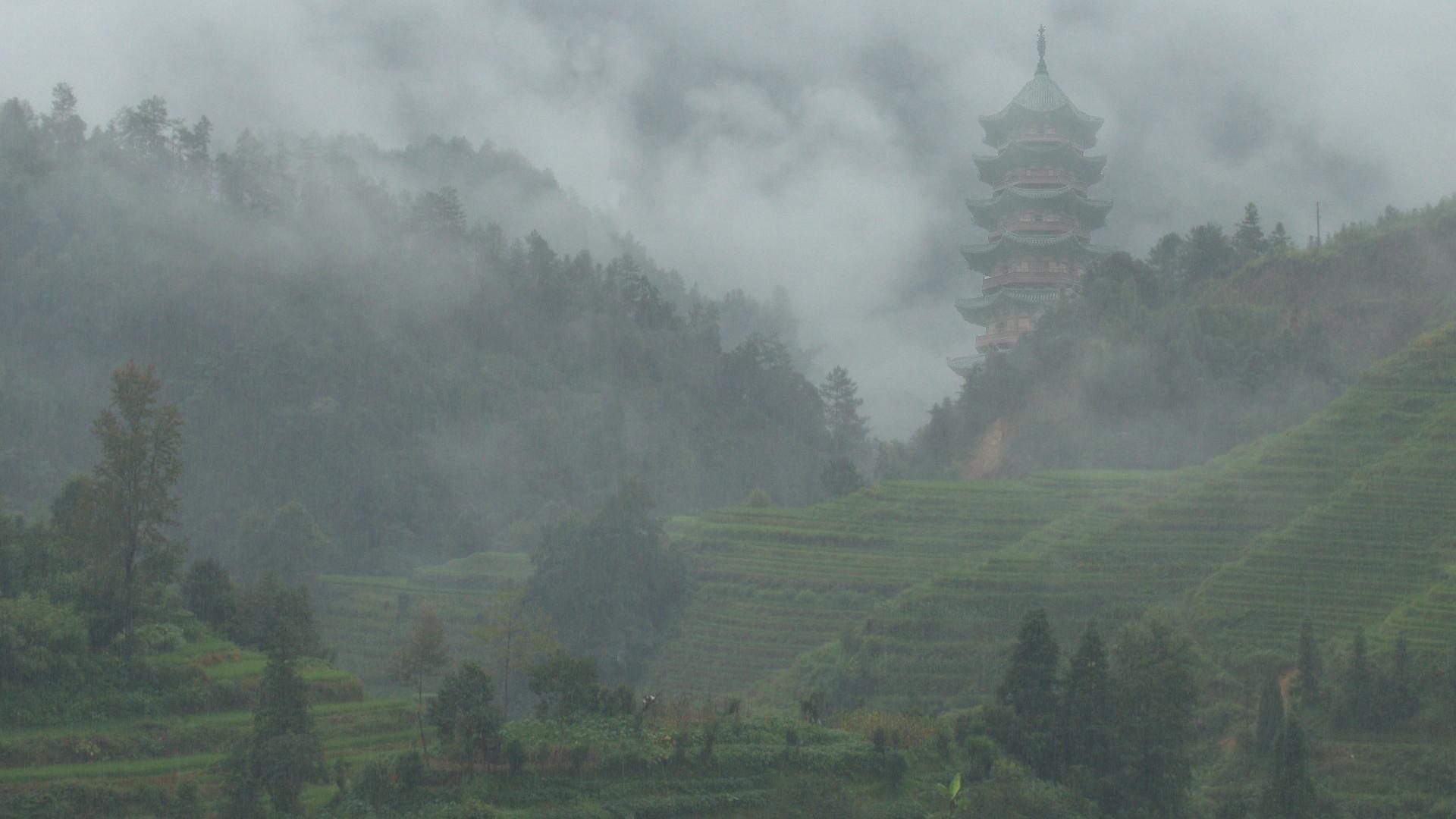}  &
		\includegraphics[width = 0.23\linewidth]{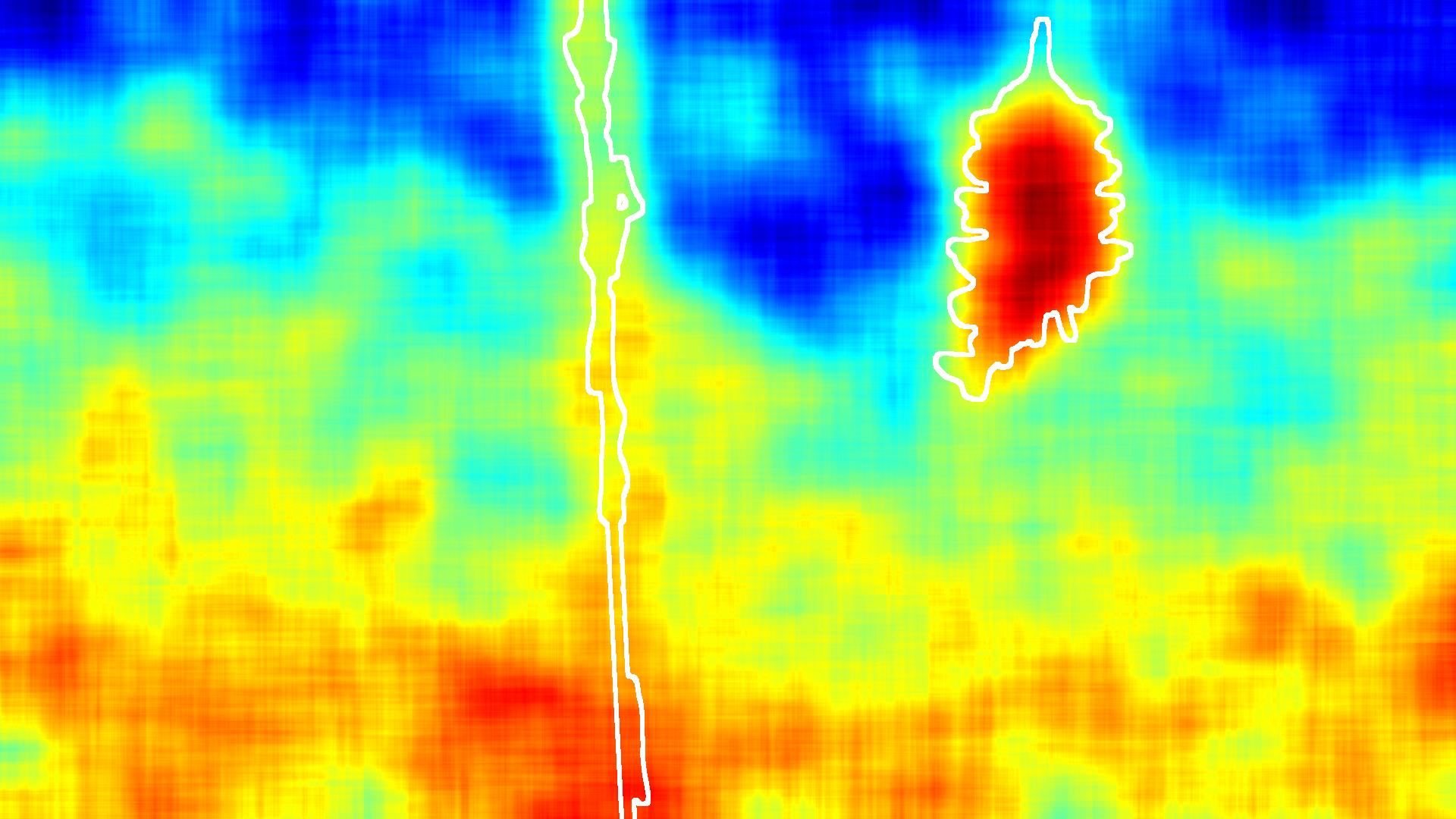} &
		\includegraphics[width = 0.23\linewidth]{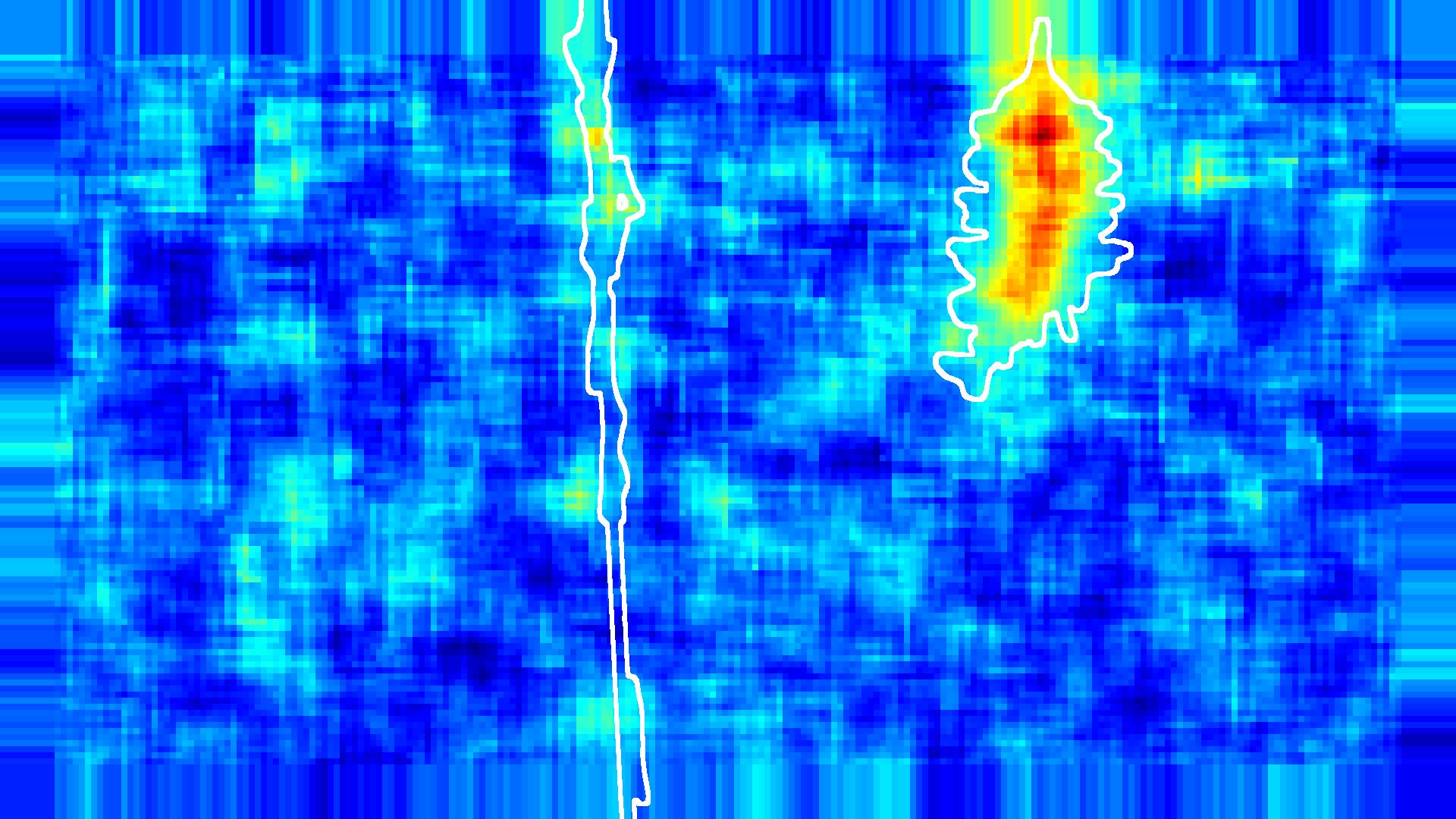} &
		\includegraphics[width = 0.23\linewidth]{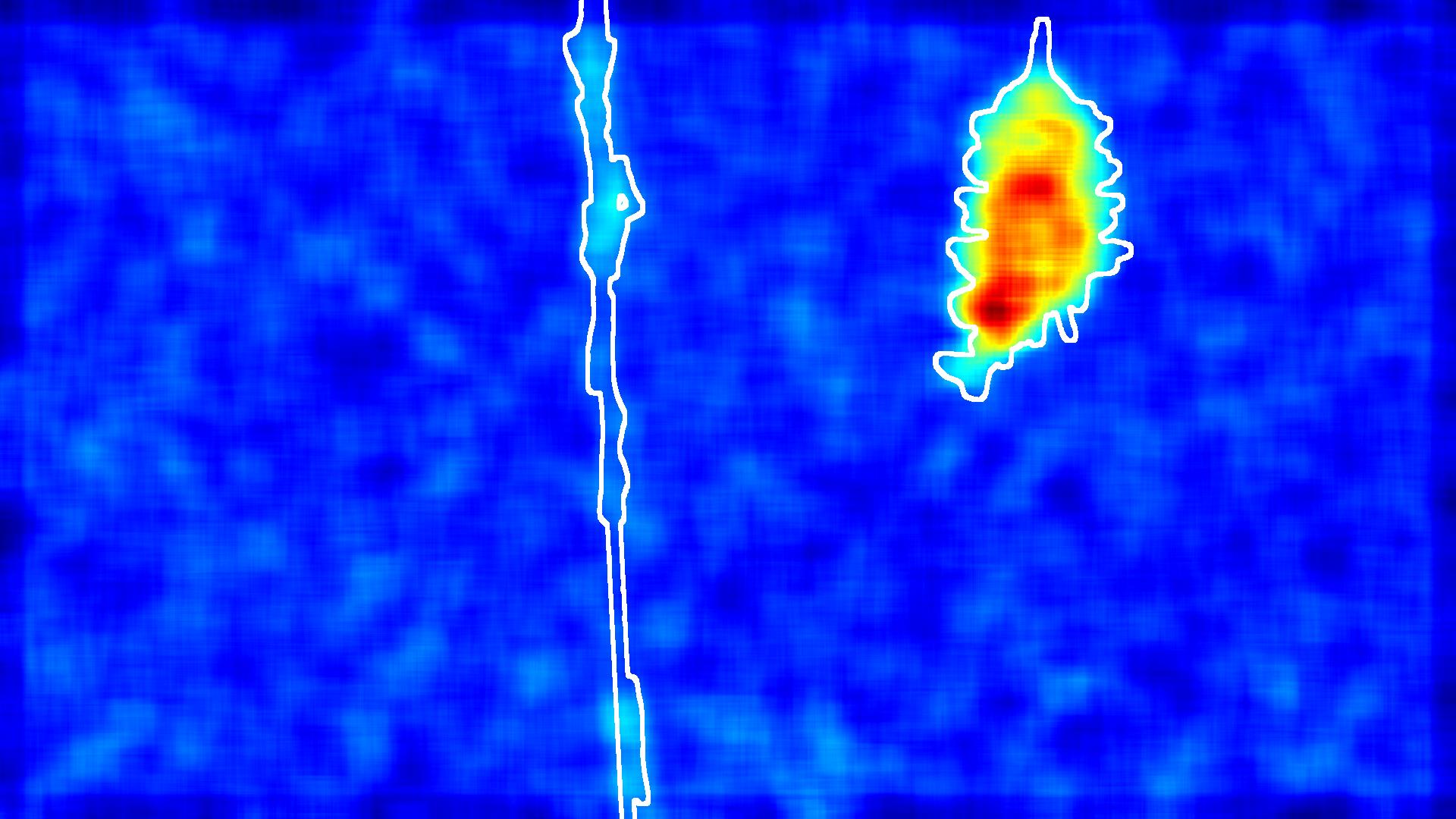} \\ \rule{0mm}{24mm} 
		\includegraphics[width = 0.23\linewidth]{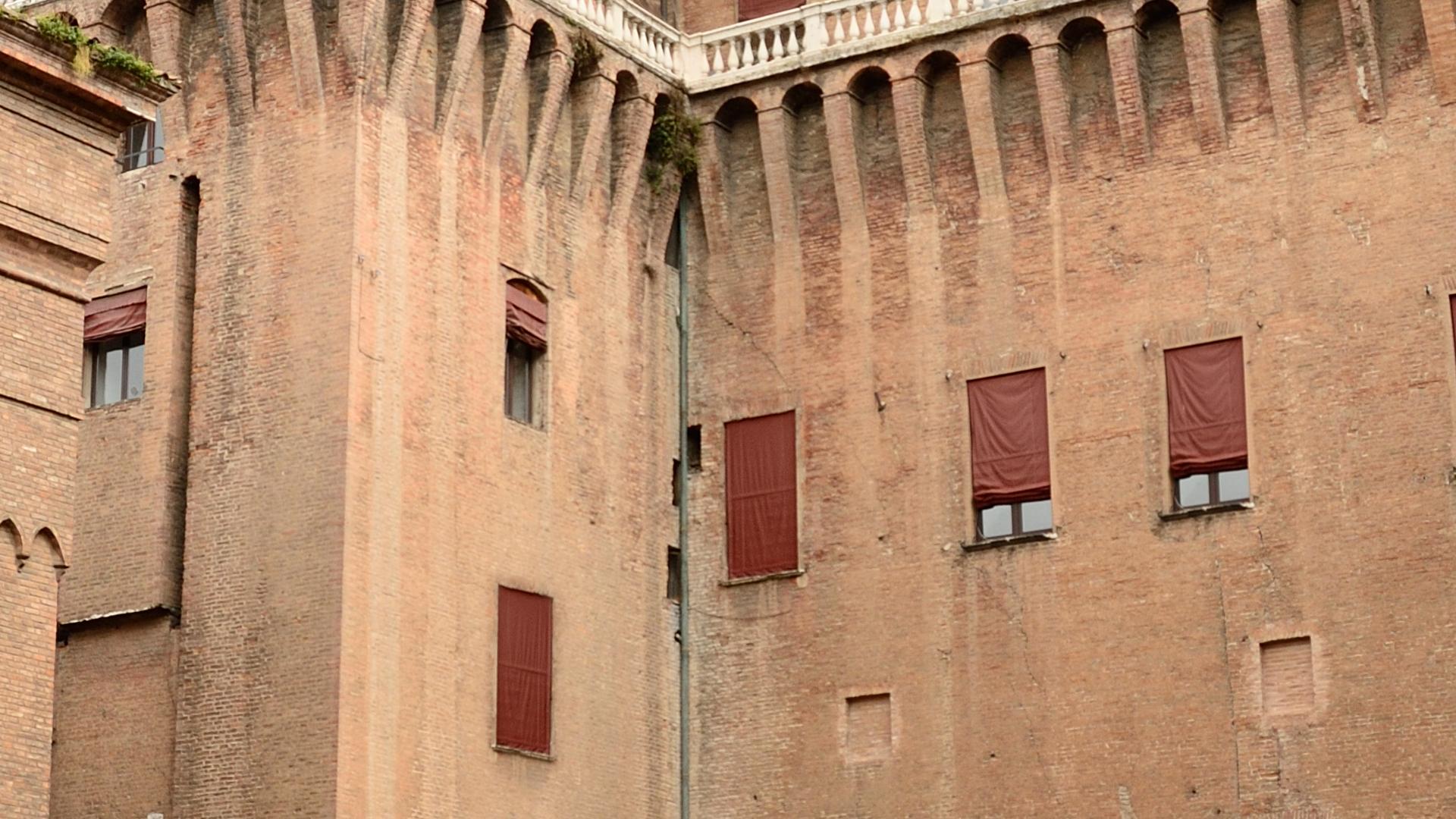}  &
		\includegraphics[width = 0.23\linewidth]{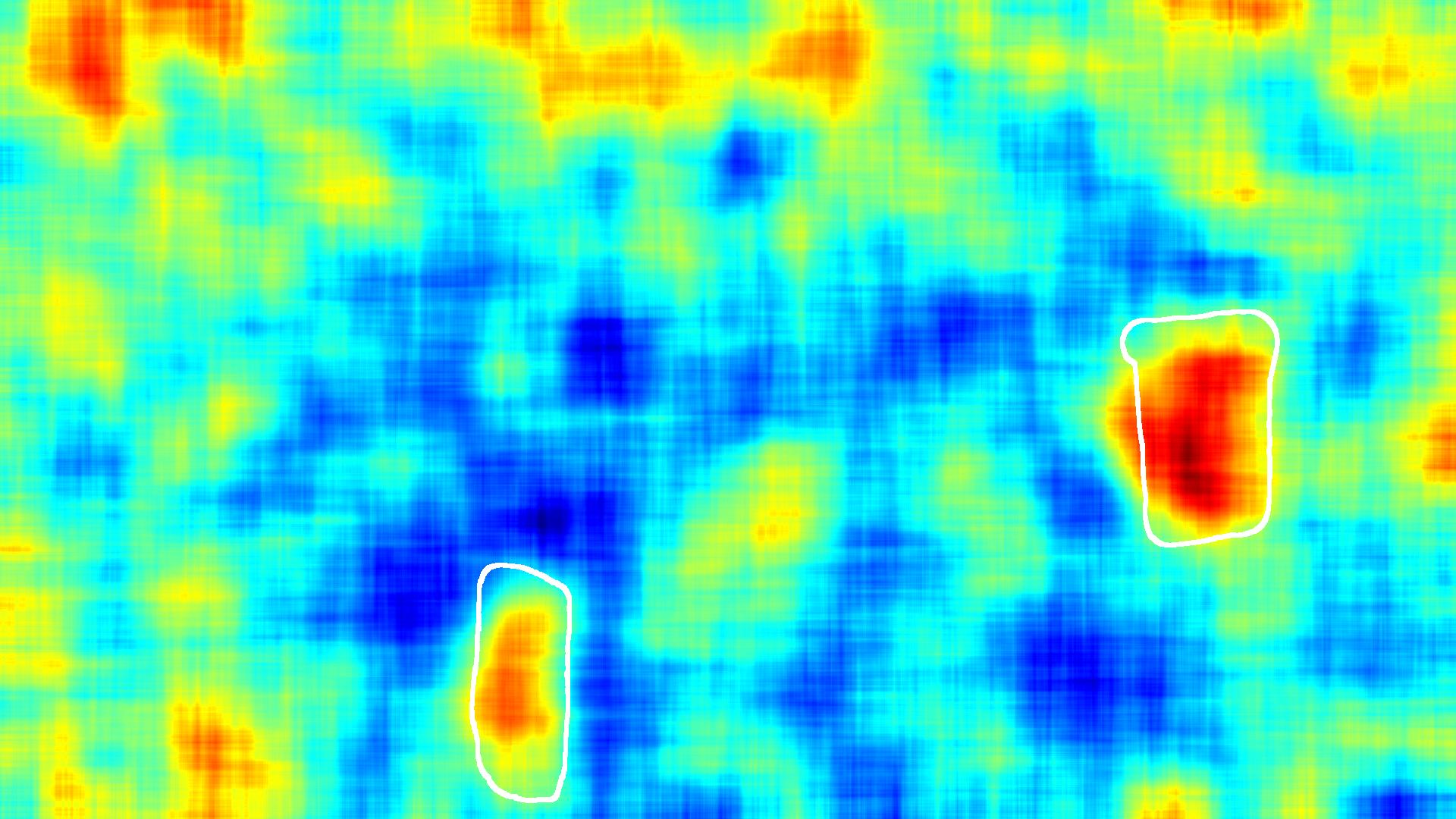} &
		\includegraphics[width = 0.23\linewidth]{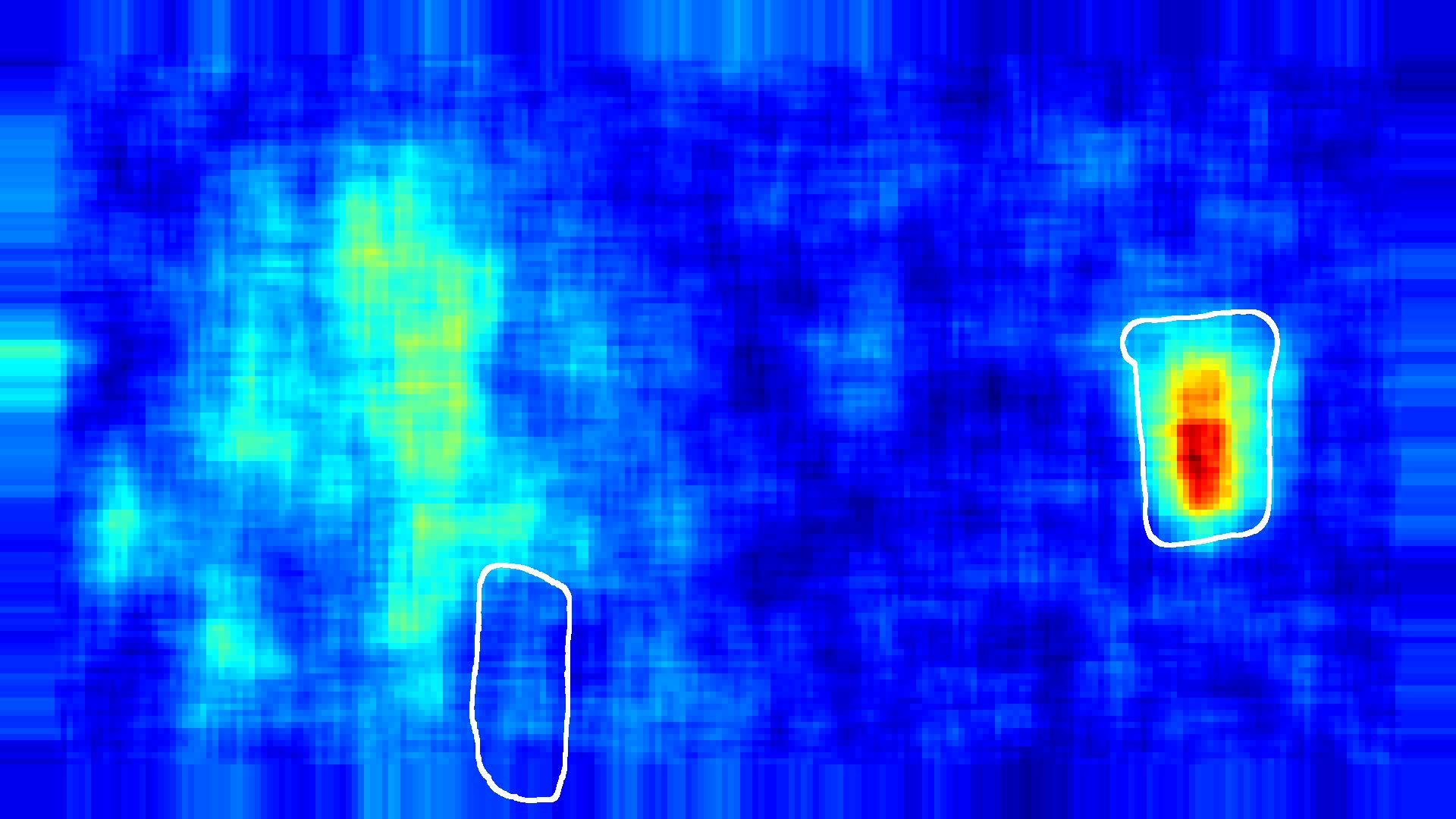} &
		\includegraphics[width = 0.23\linewidth]{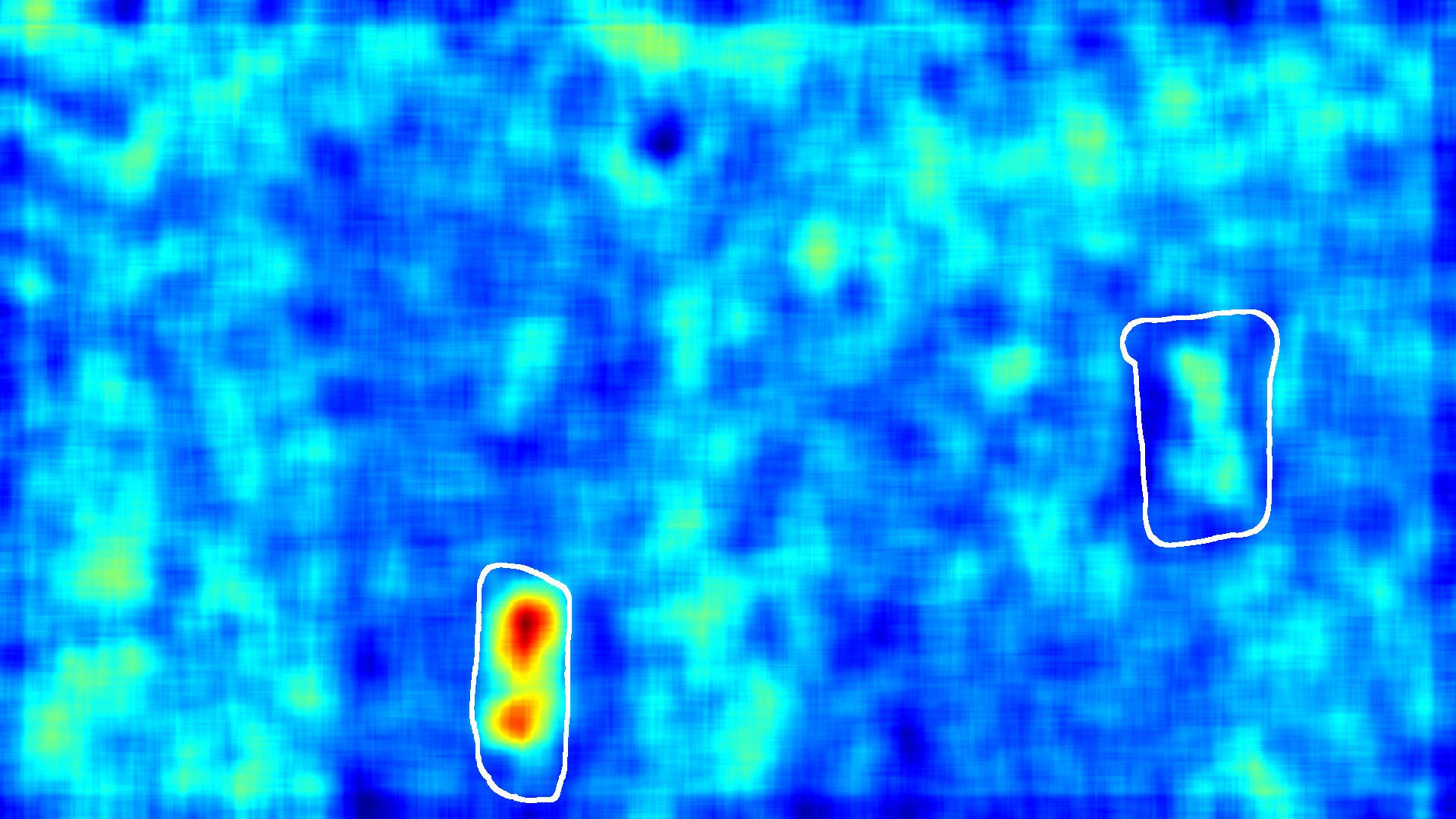} \\
		{\small (a) Forged image} &
		{\small (b) Lukas2006} &
		{\small (c) Verdoliva2014} &
		{\small (d) Proposed} \\
	\end{tabular}
	\caption{Forgery localization results for some selected examples using the dataset proposed in \cite{Korus2017}.
		Even if the images are in raw format and estimation is carried out on less data with respect to PRNU, results for the proposed approach are promising.}
	\label{fig:ex_con}
\end{figure*}


\section{Conclusions}

We performed image forgery localization by a sliding-window comparison of image residual and camera fingerprint.
Unlike in the recent literature, we do not use the PRNU fingerprint to this end, 
but a new camera {\em model} fingerprint, called noiseprint, extracted by means of a suitably trained Siamese CNN.
Experiments show that the noiseprint-based method outperforms largely the PRNU-based baseline,
and keeps providing very good results even when the fingerprint is estimated on a very small number of images.
Overall, noiseprint appears to have a great potential for multimedia forensic analyses, both in supervised and blind settings.
\balance
\bibliographystyle{IEEEtran}
\bibliography{refs}

\end{document}